\documentclass[journal]{IEEEtran}
\IEEEoverridecommandlockouts

\def\BibTeX{{\rm B\kern-.05em{\sc i\kern-.025em b}\kern-.08em
    T\kern-.1667em\lower.7ex\hbox{E}\kern-.125emX}}

\usepackage{cite}
\usepackage[utf8]{inputenc}
\usepackage{graphicx}
\usepackage{listings}
\usepackage[dvipsnames]{xcolor}
\usepackage{float}
\usepackage{booktabs}
\usepackage{algorithmic, algorithm, }
\usepackage{threeparttable}
\usepackage{amsmath, amsfonts, amsthm, amssymb, mathtools, xparse, stackengine}
\usepackage{footnote}
\makesavenoteenv{threeparttable}
\usepackage{tablefootnote}
\stackMath

\definecolor{custom_blue}{HTML}{1F77B4}
\definecolor{custom_pink}{HTML}{E377C2}
\definecolor{custom_orange}{HTML}{FF7F0E}
\definecolor{custom_purple}{HTML}{9467BD}
\definecolor{custom_green}{HTML}{2CA02C}
\definecolor{custom_red}{HTML}{D62728}
\definecolor{custom_brown}{HTML}{8C564B}

\usepackage{tikz}
\usepackage{pgfplots}
\pgfplotsset{compat=1.17}

\newcommand{\blue}{\raisebox{2pt}{\tikz{\draw[custom_blue, solid, line width=2.3pt](0,0) -- (5mm,0);}}}

\newcommand{\orange}{\raisebox{2pt}{\tikz{\draw[custom_orange, solid, line width=2.3pt](0,0) -- (5mm,0);}}}

\newcommand{\green}{\raisebox{2pt}{\tikz{\draw[custom_green, solid, line width=2.3pt](0,0) -- (5mm,0);}}}
\newcommand{\red}{\raisebox{2pt}{\tikz{\draw[custom_red, solid, line width=2.3pt](0,0) -- (5mm,0);}}}
\newcommand{\brown}{\raisebox{2pt}{\tikz{\draw[custom_brown, solid, line width=2.3pt](0,0) -- (5mm,0);}}}

\usepackage{hyperref}

\usepackage[utf8]{inputenc}
\usepackage[margin=1in]{geometry}

\usepackage{graphicx,float}
\usepackage{subfigure}

\usepackage{xcolor}

\definecolor{codegreen}{rgb}{0,0.6,0}
\definecolor{codegray}{rgb}{0.5,0.5,0.5}
\definecolor{codepurple}{rgb}{0.58,0,0.82}
\definecolor{backcolour}{rgb}{0.95,0.95,0.92}

\lstdefinestyle{mystyle}{
  backgroundcolor=\color{backcolour},   commentstyle=\color{codegreen},
  keywordstyle=\color{magenta},
  numberstyle=\tiny\color{codegray},
  stringstyle=\color{codepurple},
  basicstyle=\ttfamily,
  breakatwhitespace=false,         
  breaklines=true,                 
  captionpos=b,                    
  keepspaces=true,                 
  numbers=left,                    
  numbersep=5pt,                  
  showspaces=false,                
  showstringspaces=false,
  showtabs=false,                  
  tabsize=2
}

\newtheorem{remark}{Remark}

\theoremstyle{definition}

\begin{document}

\title{Parameter-free Reduction of the Estimation Bias in Deep Reinforcement Learning for Deterministic Policy Gradients}

\author{Baturay~Saglam,
        Furkan~Burak~Mutlu,
        Dogan~Can~Cicek,
        and~Suleyman~Serdar~Kozat,~\IEEEmembership{Senior~Member,~IEEE}%
        }
\maketitle

\begin{abstract}
Approximation of the value functions in value-based deep reinforcement learning induces overestimation bias, resulting in suboptimal policies. We show that when the reinforcement signals received by the agents have a high variance, deep actor-critic approaches that overcome the overestimation bias lead to a substantial underestimation bias. We first address the detrimental issues in the existing approaches that aim to overcome such underestimation error. Then, through extensive statistical analysis, we introduce a novel, parameter-free Deep Q-learning variant to reduce this underestimation bias in deterministic policy gradients. By sampling the weights of a linear combination of two approximate critics from a highly shrunk estimation bias interval, our Q-value update rule is not affected by the variance of the rewards received by the agents throughout learning. We test the performance of the introduced improvement on a set of MuJoCo and Box2D continuous control tasks and demonstrate that it considerably outperforms the existing approaches and improves the state-of-the-art by a significant margin.
\end{abstract}


\section{Introduction}
The policy optimization in reinforcement learning (RL) has achieved notable successes in a wide range of sequential decision-making tasks, such as for robotic control \cite{mach_5, mach_6} or time-series prediction \cite{mach_7}. However, in the deep setting of RL, where deep neural networks approximate value functions and policies, there exist several issues \cite{td3}. The systematic estimation bias that prevents the learning agents from attaining maximum performance and applicability of the deep techniques to diverse real-world tasks is one of the difficulties originating from the function approximation \cite{sutton88, td3}. For discrete action spaces, the estimation bias on the value estimates has been widely investigated for the value-based RL algorithms \cite{ddqn, max_min, precup_2001, impala, munos_safe_and_efficient_off_policy}. In addition, similar work is done in the continuous action domains with actor-critic techniques for the subtype of the estimation bias, namely, overestimation bias \cite{td3}. However, we demonstrated in our recent work \cite{tcd3} that actor-critic methods that overcome the overestimation bias and accumulated high variance induce an underestimation bias on the action-value estimates.

In continuous control, the estimation bias on the action-value estimates is generally examined under underestimation and overestimation \cite{tcd3}. Overestimation bias, caused by the maximization of the noisy estimates in traditional Q-learning \cite{q_learning}, results in a cumulative estimation error on the action values (state-action values or Q-values) throughout the learning stage \cite{td3}. As deep neural networks represent the action and value functions in the deep RL setting, such a function approximation noise is inevitable \cite{td3}. Due to the temporal difference (TD) learning \cite{sutton88}, this inaccuracy in the value estimation is further amplified \cite{td3}. The underestimation bias, in contrast, is an outcome of Q-learning \cite{q_learning} variants that focus on eliminating the accumulated overestimation bias \cite{tcd3}. Although a recent objective function proposal in the Twin Delayed Deep Deterministic Policy Gradient (TD3) algorithm \cite{td3}, Clipped Double Q-learning, is shown to eliminate the overestimation bias and accumulated variance, it can nevertheless decrease an RL agent’s performance by assigning low values to optimal state-action pairs and thus, may result in suboptimal policies and divergent behaviors \cite{tcd3}.

In the Clipped Double Q-learning algorithm \cite{td3}, two Q-networks with identical structures and different parameters are initialized before the learning process \cite{td3}. The minimum of these critics' estimates is utilized to form the objective of Q-networks during learning. Despite the decoupled actor and critics, using the minimum Q-value in learning the targets results in persistent underestimation of the state-action values \cite{tcd3}. Recent works, Weighted Delayed Deep Deterministic Policy Gradient (WD3) \cite{wd3} and Triplet-average Deep Deterministic Policy Gradient (TADD) \cite{tadd}, focus on this existing underestimation bias in the TD3 algorithm \cite{td3} and introduce a linear combination of the functions of action-value estimates in forming the objective of Q-networks. Although the recent objective function proposals \cite{wd3} and \cite{tadd} are shown to reduce the underestimation bias and improve the state-of-the-art, their theoretical assumptions on the underestimation of Q-values are either on a strong basis or infeasible assumptions that prevent their approach to be adapted to the off-policy learning. Furthermore, our recent work for the problem of the underestimation bias, Triplet Critic Deep Deterministic Policy Gradient (TCD3) \cite{tcd3}, heuristically searches for an alternative for the Q-network objective and proposes a combination of three approximate critics. However, maintaining three deep networks comes with a great computational complexity compared to the TD3 algorithm \cite{td3}. 

In this paper, we extend our previous study \cite{tcd3} on the estimation bias such that we first examine the current strategies that aim to overcome the underestimation bias in deterministic policy gradient (DPG) \cite{dpg} algorithms. We address the detrimental issues in these algorithms and explain the infeasible assumptions in their theoretical background. We then derive a closed-form expression for the estimation error yielded by the Clipped Deep Q-learning algorithm \cite{td3} and our previous work TCD3 \cite{tcd3}, without any statistical assumptions that violate the off-policy RL paradigm, which was not introduced in our previous work. Using the derived closed-form expressions, we introduce a new variant of Deep Q-learning \cite{dqn}, Stochastic Weighted Twin Critic Update, that achieves superior performance to our previous work but with using only two critics and hence, having a 33\% less time complexity. Our approach derives a parameter-free linear combination of the functions of two approximate critics, weights of which are sampled from a bias interval that corresponds to a significantly smaller underestimation bias than the existing approaches. In addition to our previous work on the underestimation bias, the main contributions of this study can be summarized as follows: 
\begin{itemize}
    \item We first address the issues with the existing algorithms that focus on the underestimation in deterministic policy gradient \cite{dpg} methods. We explain why the statistical assumptions made in those works cannot be adapted to the off-policy deep RL in continuous action spaces. 
    \item We derive a closed-form expression for the estimation bias in the Clipped Double Q-learning algorithm \cite{td3} and TCD3 \cite{tcd3}, and theoretically show that if the rewards that the agent receives vary on a large scale, the underestimation of the action-value estimates detrimentally increases. 
    \item Through an extensive statistical analysis of the expected error in the existing approaches and derived closed-form expressions, we introduce a stochastic Q-network update rule in which weights are sampled from a bias interval that is substantially smaller than the expected errors in the existing approaches and TCD3 \cite{tcd3}. 
    \item We empirically verify our claims by comparing the actual and estimated Q-values produced by the WD3 \cite{wd3} and TADD \cite{tadd} algorithms and demonstrating that an increasing variance of the received reward signals increases the underestimation throughout the learning.
    \item Our method is not affected by the variance of the reward signals as it samples the weights of the Q-networks from an interval, the lower bound of which is linearly decreased. An extensive set of empirical studies in several challenging OpenAI Gym \cite{gym} tasks reflect our theoretical claims and show that the introduced approach dramatically outperforms the competing methods and improves our previous study.  
    \item The source code of our algorithm is publicly available at our GitHub repository\footnote{\url{https://github.com/baturaysaglam/SWTD3}\label{our_repo}} to ensure reproducibility.
\end{itemize}

\section{Related Work}
Prior studies on the approximation error in reinforcement learning have been done by \cite{mach_3, mach_4} in terms of the estimation bias and resulting high variance build-up. This paper focuses on one of the function approximation error outcomes, namely, underestimating the action-values. In the following, we extensively investigate the background of the estimation error phenomenon in deep reinforcement learning. 

\subsection{Discrete Action Spaces}
The estimation error induced by the maximization of Q-values has been extensively studied in discrete action spaces. For Deep Q-learning \cite{dqn}, many techniques were proposed to mitigate the impacts of the overestimation bias caused by the function approximation and policy optimization. Van Hasselt et al. \cite{ddqn} address the function approximation error for discrete action spaces in their work, Deep Double Q-learning (DDQN), which is one of the successor studies to the Deep Q-learning \cite{dqn}. By employing two independent and identically structured Q-value approximators, DDQN \cite{ddqn} obtains unbiased Q-value estimates. Lan et al. \cite{max_min} modifies Deep Q-learning \cite{dqn} through the utilization of multiple action-value estimators. Their approach, Maxmin Q-learning \cite{max_min}, uses multiple action-value estimates selected through partial maximum operators, the minimum of which constructs the Deep Q-learning target \cite{dqn}. Although \cite{max_min} primarily aims to eliminate the overestimation, they show that their method may yield in underestimation \cite{max_min}. Additionally, methods that employ multi-step returns are shown to overcome the estimation bias \cite{impala,precup_2001,munos_safe_and_efficient_off_policy,weighted_q_learning, mach_2} and proven to be effective through distributed approaches \cite{impala}, weighted Q-learning \cite{weighted_q_learning}, and importance sampling for off-policy correction \cite{munos_safe_and_efficient_off_policy, precup_2001, impala,trust_region_estimators}. However, these methods introduce a trade-off between the biased action-value estimates and accumulated variance, as shown by \cite{trust_region_estimators}.
Furthermore, these approaches use impractical longer horizons than one-step solutions to the bias-variance trade-off \cite{trust_region_estimators}. Moreover, \cite{petrik} proposed a one-step improvement for the reduction of the contribution of each erroneous estimate by reducing the discount factor in a structured manner.

\subsection{Continuous Action Spaces}
Although the estimation bias in discrete action space is overcome by the existing Deep Q-learning \cite{dqn} variants, they cannot be adapted to the control of the continuous systems due to the employment of a separate actor network \cite{td3}. As there exist infinitely many intractable actions in continuous action domains, the maximization of Q-networks cannot be used to select actions. Hence, the previously introduced methods for discrete action domains cannot be used \cite{td3}. Nonetheless, a direct and one-step solution to the overestimation and variance accumulation has been proposed by \cite{td3}. It is shown to be effective in eliminating the function approximation error for the deep setting of RL. Their research demonstrates that the deep function approximation of Q-values causes overestimation bias and cumulative variance in continuous action domains, which causes the approximate gradient of the actor network to diverge from the actual gradient. An extension of temporal difference learning \cite{precup_2001} in the DPG \cite{dpg} methods, Clipped Double Q-learning \cite{td3}, on which we build our algorithm, presents a direct remedy to the overestimation problem by employing two identically structured Q-networks. On top of Clipped Double Q-learning \cite{td3}, the delayed actor updates and target policy regularization through additive policy noise constitute the TD3 algorithm \cite{td3}, which is shown to exhibit a state-of-the-art performance. TD3 \cite{td3} overcomes the overestimation build-up by performing the target Q-value computation through the minimum of two approximate critics. Their introduced update rule, Clipped Double Q-learning \cite{td3}, is used in many state-of-the-art continuous control algorithms.

Although the improvements introduced by Clipped Double Q-learning \cite{td3} can eliminate cumulative estimation error, the use of the minimum of two critics causes an underestimation bias in the Q-value estimations, as stated by \cite{td3} and empirically shown by \cite{tcd3,wd3,tadd}. Several techniques have been proposed to address the underestimation problem, including the use of a linear combination of the Q-value estimates by approximate critics to compute the objective for Q-network update \cite{wd3, tadd}. We extensively review these approaches along with our previous work TCD3 \cite{tcd3} in the later sections. 

\section{Background}
Reinforcement learning paradigm considers an agent that interacts with its environment to learn the optimal, reward-maximizing behavior. The standard reinforcement learning is represented by a partially or fully observable Markov Decision Process (MDP) defined by the tuple $(\mathcal{S}, \mathcal{A}, p, \gamma)$ where $\mathcal{S}$ and $\mathcal{A}$ denote the state and action spaces, respectively, $p$ is the transition dynamics and $\gamma$ is the constant discount factor. At each discrete time step $t$, the agent observes its state $s \in \mathcal{S}$ and chooses an action $a \in \mathcal{A}$ according to its policy $\pi_{\phi}$, stochastic or deterministic, parameterized by $\phi$. Then, based on its action decision given the observed state, the agent receives the reward $r$ from a reward function corresponding to its environment, and observes a next state $s'$ such that $s', r \sim p(s, a)$. The objective of the agent is to maximize the cumulative reward defined as the discounted sum of future rewards $R_{t} = \sum_{i = t}^{T}\gamma^{i - t}r(s_{i}, a_{i})$ where the discount factor $\gamma \in [0, 1)$ downscales the long-term rewards to prioritize the short-term rewards more.  

The agent learns the optimal policy $\pi^{*}$ that maximizes the expected return $\mathbb{E}_{s_{i} \sim p_{\pi}, a_{i} \sim \pi}[R_{0}]$. In actor-critic settings where action space is continuous, parameterized policies $\pi_{\phi}$ represented by deep neural networks with parameters $\phi$ are optimized by computing the gradient of the expected return $\nabla_{\phi}J(\phi)$ through a policy gradient technique. In this study, we consider the deterministic policy gradient algorithm expressed by:
\begin{equation}
    \nabla_{\phi}J(\phi) = \mathbb{E}_{s \sim p_{\pi}}[\nabla_{a}Q^{\pi}(s, a) \vert_{a = \pi(s)} \nabla_{\phi}\pi_{\phi}(s)].
\end{equation}
The expected return after taking the action $a$ given the observed state $s$ under the current policy $\pi$ is computed by the critic (action-value function or Q-function) $Q^{\pi}(s, a) = \mathbb{E}_{s_{i} \sim p_{\pi}, a_{i} \sim \pi}[R_{t} \vert s, a]$ which values the quality of the action decision given the observed state while following the current policy $\pi$. The critic evaluates and improves the agent's policy to obtain higher quality action choices.

In standard Q-learning \cite{q_learning}, if the transitions dynamics of the environment is accessible, the action-value function $Q^{\pi}$ is estimated through recursive Bellman optimization \cite{belmann} given the transition tuple $(s, a, r, s')$:
\begin{equation}
    Q^{\pi}(s, a) = r + \gamma\mathbb{E}_{s', a'}[Q^{\pi}(s', a')]; \quad a' \sim \pi(s').
\end{equation}

For large state and action spaces, the action-value function is usually estimated by function approximators $Q_{\theta}(s, a)$ parameterized by $\theta$, also known as the Q-networks. In the deep setting of Q-learning \cite{q_learning}, the Q-network is updated through the temporal difference learning \cite{precup_2001} by a secondary frozen target network $Q'_{\theta}(s, a)$ to construct the objective for behavioral Q-network, also known as Deep Q-learning \cite{dqn}:
\begin{equation}
    y = r + \gamma Q_{\theta'}(s', a'); \quad a' \sim \pi_{\phi'}(s'),
\end{equation}
where the next actions given the observed next state can be obtained from a separate target actor network $\pi_{\phi'}$ for actor-critic settings in continuous control. The target networks are either updated by a small proportion $\tau$ at each time step, i.e., $\theta' \leftarrow \tau\theta + (1 - \tau)\theta'$, called soft-update, or periodically to exactly match the behavioral networks called hard-update.

\section{The Underestimation Bias in Deterministic Policy Gradients}
\subsection{An Informative Analysis on the Existing Approaches to the Underestimation Bias}
We start by explaining the current approaches to the underestimation bias in the literature. Mainly, we investigate the WD3 \cite{wd3} and TADD \cite{tadd} algorithms and the theoretical background of these algorithms. These studies extend the Clipped Double Q-learning algorithm \cite{td3} by replacing the Q-networks' objective with a fixed linear combination, as discussed. Let us first consider the WD3 algorithm \cite{wd3}. In the simplest terms, Q-networks are updated as follows:
\begin{gather}
\label{eq:wd3_upt}
\text{\small$
    y = r + \gamma \left(\beta\underset{i = 1, 2}{\mathrm{min}} Q_{\theta'_{i}}(s', \Tilde{a}') + \frac{1 - \beta}{2}\sum_{i = 1}^{2}Q_{\theta'_{i}}(s', \Tilde{a}')\right),$} \\
    J(\theta_{i}) = \|y - Q_{\theta_{i}}(s, a)\|^{2},
\end{gather}
where $\Tilde{a}'$ is the action chosen by the target policy in the next state $s'$ perturbed by a zero-mean Gaussian noise, i.e., $\Tilde{a}' = \pi_{\phi'}(s') + \mathcal{N}(0, \sigma)$, $\sigma$ is the standard deviation of the perturbation noise, and $J(\theta_{i})$ is the loss associated with critic $Q_{\theta_{i}}$. Here, $\beta \in [0, 1]$ is a parameter that controls the underestimation since minimum operator yields the underestimation of Q-values \cite{td3, tcd3}. Note that this additive exploratory noise does not alter the expected function approximation error by having a zero mean \cite{td3}.

The TADD algorithm \cite{tadd} adopts a similar approach through an additional third critic employed in Q-learning \cite{q_learning}. In addition, the last $K$ parameters of the third critic is stored in a critic network buffer, which is used to construct the objective for the Q-networks, particularly:
\begin{gather}
\label{eq:tadd_upt}
\text{\small$
    y = r + \gamma\left(\beta\underset{i = 1, 2}{\mathrm{min}} Q_{\theta'_{i}}(s', \Tilde{a}') + \frac{(1 - \beta)}{K}\sum_{k}Q_{\theta'_{3,k}}(s', \Tilde{a}')\right),$} \\
    J(\theta_{i}) = \|y - Q_{\theta_{i}}(s, a)\|^{2},
\end{gather}
where the average of last $K$ assists in reducing the variance of the Q-value estimates \cite{tadd}. 

In these studies, the errors by the employed Q-networks are represented by probability distributions, which is feasible as the employment of deep neural networks and bootstrapping in the Q-learning introduce noise in the action-value estimates \cite{td3,sutton88}. Based on such a probabilistic representation, there exist two assumptions made by these works based on the error distributions. First, it is stated that the error by each of the critics can be represented either by a zero-mean Gaussian or a zero-mean uniform distribution. Second, error distributions by the two critics are independent and identically distributed, as shown by Theorem 1 and 2 in \cite{wd3} and by Theorem 1 in \cite{tadd}. Formally, we express the made assumptions in \cite{wd3} and \cite{tadd} as:
\begin{gather}
    N_{i} \sim \mathcal{N}(0, \Tilde{\sigma}), \quad
    Z_{i} \sim \mathrm{uniform}[-\delta, \delta]; \\
    \left(Q_{\theta_{i}}(s, a) - Q^{*}(s, a)\right) \sim N_{i}\vee Z_{i}:\\
    \mathrm{P}(N_{1} \cap N_{2}) = \mathrm{P}(N_{1})P(N_{2}), \\
    \mathrm{P}(Z_{1} \cap Z_{2}) = \mathrm{P}(Z_{1})P(Z_{2}),
\end{gather}
for some parameters $\delta$ and $\Tilde{\sigma}$, where $Q^{*}$ denotes the actual Q-value of the state-action pair $(s, a)$. However, the zero-mean assumption violates the existence of the estimation bias:
\begin{gather}
    \mathbb{E}[Q_{\theta_{i}}(s, a) - Q^{*}(s, a)] = 0, \\
    \mathbb{E}[Q_{\theta_{i}}(s, a)] = Q^{*}(s, a).\label{eq:unbiased_est}
\end{gather}
The latter equation is satisfied since $Q^{*}(s, a)$ is the fixed point of the Bellman operator $\mathcal{T}^{\pi^{*}}$ \cite{belmann} under the optimal policy $\pi^{*}$ \cite{munos_safe_and_efficient_off_policy}. Then, from (\ref{eq:unbiased_est}), we infer that each $Q_{\theta_{i}}(s, a)$ is an unbiased estimator of $Q^{*}(s, a)$ which contradicts with the existence of an estimation bias. Furthermore, errors of the two critics cannot be entirely independent due to the employment of the opposite critic in learning the targets, as well as the same replay buffer \cite{td3}. Therefore, assumptions made in the current approaches to the underestimation bias violate the nature of the Q-learning in off-policy and deterministic policy gradient \cite{dpg} methods. Finally, we can conclude this section with the following remarks. 

\begin{remark}
\label{rem:zero_mean}
    Estimation error by the two critics in the Clipped Double Q-learning algorithm \cite{td3} cannot follow a zero-mean probability distribution. If so, then the existence of an estimation bias is violated.
\end{remark}

\begin{remark}
\label{rem:iid}
    Error distributions by the two critics in the Clipped Double Q-learning algorithm \cite{td3} are not independent due to the employment of the opposite critic in learning the targets and the use of the same replay buffer.
\end{remark}

\subsection{Derivation of the Closed-Form Expression for the Underestimation Bias}
By considering Remark \ref{rem:zero_mean} and \ref{rem:iid}, we begin to derive a closed-form expression for the estimation bias in the Clipped Double Q-learning algorithm \cite{td3}. The presence and effects of overestimation in actor-critic settings are highlighted in \cite{td3} through the gradient ascent in the policy updates. However, using the minimum operator to compensate for the overestimation of Q-values may result in underestimated action-value estimates. We begin by proving through basic assumptions and claims that the underestimation phenomenon exists in DPG \cite{dpg} algorithms for environments with varying reinforcement signals. We follow the gradient ascent approach in \cite{td3} to show such underestimation.

In the TD3 algorithm \cite{td3}, the policy is updated using the minimum value estimate by two approximate critics, $Q_{\theta_{1}}$ and $Q_{\theta_{2}}$, parameterized by $\theta_{1}$ and $\theta_{2}$, respectively. Without loss of generality, we assume that both critics overestimate the action-values, and the policy is updated with respect to the first approximate critic $Q_{\theta_{1}}(s, a)$ through the DPG algorithm \cite{dpg}. The assumption on the overestimation of both Q-networks is valid as the single Q-network in the Deep Deterministic Policy Gradient (DDPG) algorithm \cite{ddpg} already overestimates the Q-values, as shown by \cite{td3}. First, let $\phi_{\mathrm{approx}}$ define the parameters from the actor update by the maximization of the first approximate critic $Q_{\theta_{1}}(s, a)$:
\begin{equation}
    \phi_{\mathrm{approx}} = \phi + \frac{\eta}{Z_{1}}\mathbb{E}_{s \sim p_{\pi}}[\nabla_{\phi}\pi_{\phi}(s)\nabla_{a}Q_{\theta_{1}}(s, a)\vert_{a = \pi_{\phi}(s)}],
\end{equation}
where $Z_{1}$ is the gradient normalizing term such that $Z^{-1}\|\mathbb{E}[\cdot]\| = 1$, and $\eta > 0$ is the learning rate. As the actor is optimized with respect to $Q_{\theta_{1}}(s, a)$ and the gradient direction is a local maximizer, there exists $\zeta$ sufficiently small such that if $\eta < \zeta$, then the \textit{approximate} value of the policy, $\pi_{\mathrm{approx}}$, by the first critic will be bounded below by the \textit{approximate} value of the policy by the second critic:
\begin{equation}\label{eq:td3_like_underestimation}
    \mathbb{E}[Q_{\theta_{1}}(s, \pi_{\mathrm{approx}}(s))] \geq \mathbb{E}[Q_{\theta_{2}}(s, \pi_{\mathrm{approx}}(s))].
\end{equation}
Note that for the latter equation, there could be a local maximizer for which $\mathbb{E}[Q_{\theta_{2}}(s, \pi_{\mathrm{approx}}(s))] \geq \mathbb{E}[Q_{\theta_{1}}(s, \pi_{\mathrm{approx}}(s))]$. However, such a possibility can be neglected in actor-critic algorithms that utilize Clipped Double Q-learning \cite{td3} since the actor is always optimized with respect to the first critic $Q_{\theta_{1}}$ \cite{td3}. Then, we can treat the function approximation error for both critics as distinct Gaussian random variables:
\begin{align}\label{eq:error_gaussian_def}
\begin{split}
    Q_{\theta_{1}}(s, a) - Q^{*}(s, a) &= N_{1} \sim \mathcal{N}(\mu_{1}, \sigma_{1}), \\
    Q_{\theta_{2}}(s, a) - Q^{*}(s, a) &= N_{2} \sim \mathcal{N}(\mu_{2}, \sigma_{2}).
\end{split}
\end{align}
Following (\ref{eq:td3_like_underestimation}) and Remark \ref{rem:zero_mean}, we have $\mu_{1} \geq\mu_{2} \geq 0$. As the same experience replay buffer \cite{exp_replay} and opposite critics are used in learning the target Q-values and critics, error Gaussian's denoted by (\ref{eq:error_gaussian_def}) are not entirely independent according to Remark \ref{rem:iid}. Through the first moment of the minimum of two correlated Gaussian random variables \cite{nadaraj}, the expected estimation error for the Clipped Double Q-Learning algorithm \cite{td3} becomes:
\begin{equation}
\label{eq:nadaraj_min_exp}
\text{\footnotesize$
    \mathbb{E}[\underset{i=1, 2}{\mathrm{min}}\{N_{i}\}] = \mu_{2} + (\mu_{1} - \mu_{2})\Phi(\frac{\mu_{1} - \mu_{2}}{\theta}) - \theta\psi(\frac{\mu_{1} - \mu_{2}}{\theta}),$}
\end{equation}
where $\theta \coloneqq \sqrt{\sigma_{1}^{2} + \sigma_{2}^{2} - 2\rho\sigma_{1}\sigma_{2}}$, $\rho$ is the correlation coefficient between $N_{1}$ and $N_{2}$, and $\Phi(\cdot)$ and $\psi(\cdot)$ are the cumulative distribution function (CDF) and probability density function (PDF) of the standard normal distribution, respectively. Due the presence of the delayed actor updates, the mean function approximation errors by both critics are not very distant due to the decoupled actor and first critic. Hence, for simplicity, we can assume that $\mu_{1} \approx \mu_{2}$. Using this, (\ref{eq:nadaraj_min_exp}) reduces to:
\begin{equation}
\label{eq:exp_min}
    \mathbb{E}[\underset{i=1, 2}{\mathrm{min}}\{N_{i}\}] = \mu_{1} - \frac{\theta}{\sqrt{2\pi}},
\end{equation}
since $\Phi(0) = 0.5$, $\psi(0) = 1 / \sqrt{2\pi}$. Hence, if $\sigma_{1}, \sigma_{2} > \sqrt{\pi / (1 - \rho)}\mu_{1}$, then the action-value estimate will be underestimated:
\begin{equation}
    \mathbb{E}[\underset{i=1, 2}{\mathrm{min}}\{Q_{\theta_{i}}(s, a)\} - Q^{*}(s, a)] < 0.
\end{equation}

From $\sigma_{1}, \sigma_{2} > \sqrt{\pi / (1 - \rho)}\mu_{1}$ condition, if the pair of critics are highly correlated, underestimation does not exist. However, there exists a moderate correlation between the pair of critics due to the delayed policy updates which increases the underestimation possibility \cite{td3}.

Although the improvements by \cite{td3} aim to reduce the estimation error growth, the variance of the Q-values cannot be eliminated as they are adhered to the variance of the future value estimates and rewards \cite{td3}. Furthermore, the Bellman equation \cite{belmann} in function approximation settings cannot be exactly satisfied \cite{td3}, which results in erroneous Q-value estimates as a function of the actual TD-error expressed by (\ref{eq:error_gaussian_def}). Then, we can show that the variance of the value estimates increases as the agent receives reward signals that vary on a large scale due to the exploration \cite{sutton2018reinforcement}. As shown in \cite{td3}, the Q-value estimates can be expressed in terms of the expected sum of discounted future rewards:
\begin{align}
\begin{split}
Q_{\theta_{i}}(s, a) = \mathbb{E}_{s_{i} \sim p_{\pi}, a_{i} \sim \pi}[\sum_{i = t}^{T}\gamma^{i - t}r_{i}] + \mu_{i}\sum_{i = t}^{T}\gamma^{i - t}.
\end{split}
\end{align}
If the expected estimation errors by both critics are constant, varying reinforcement signals increase the variance of the Q-value estimates resulting in an increasing underestimation bias. Since an extensive exploration is a mandatory requirement for continuous action spaces \cite{sutton2018reinforcement}, the variance of the reinforcement signals usually increases throughout the learning phase. Therefore, the underestimation bias on the value estimates becomes unavoidable. Moreover, in the underestimation case, the estimation error is not accumulated due to the TD learning \cite{sutton88, precup_2001}. Thus, the underestimation bias is far preferable to the overestimated Q-values in the actor-critic setting \cite{td3}. Nevertheless, underestimated action-values may discourage agents from choosing good state-action pairs for an extended period and reinforce agents to value suboptimal state-action pairs more frequently \cite{tcd3}.

\begin{remark}
\label{rem:increasing_variance}
  A varying set of reinforcement signals increase the variance of the Q-value estimates, which results in an increasing underestimation bias.
\end{remark}

\begin{figure*}[htbp]
    \centering
    \begin{equation*}
        \text{{\blue} SWTD3} \quad \text{\textcolor{custom_blue}{$\boldsymbol{\bullet}$} True SWTD3} \qquad \text{{\green} TD3} \quad \text{\textcolor{custom_green}{$\boldsymbol{\bullet}$} True TD3}
    \end{equation*}
	\subfigure{
		\includegraphics[width=1.01in, keepaspectratio]{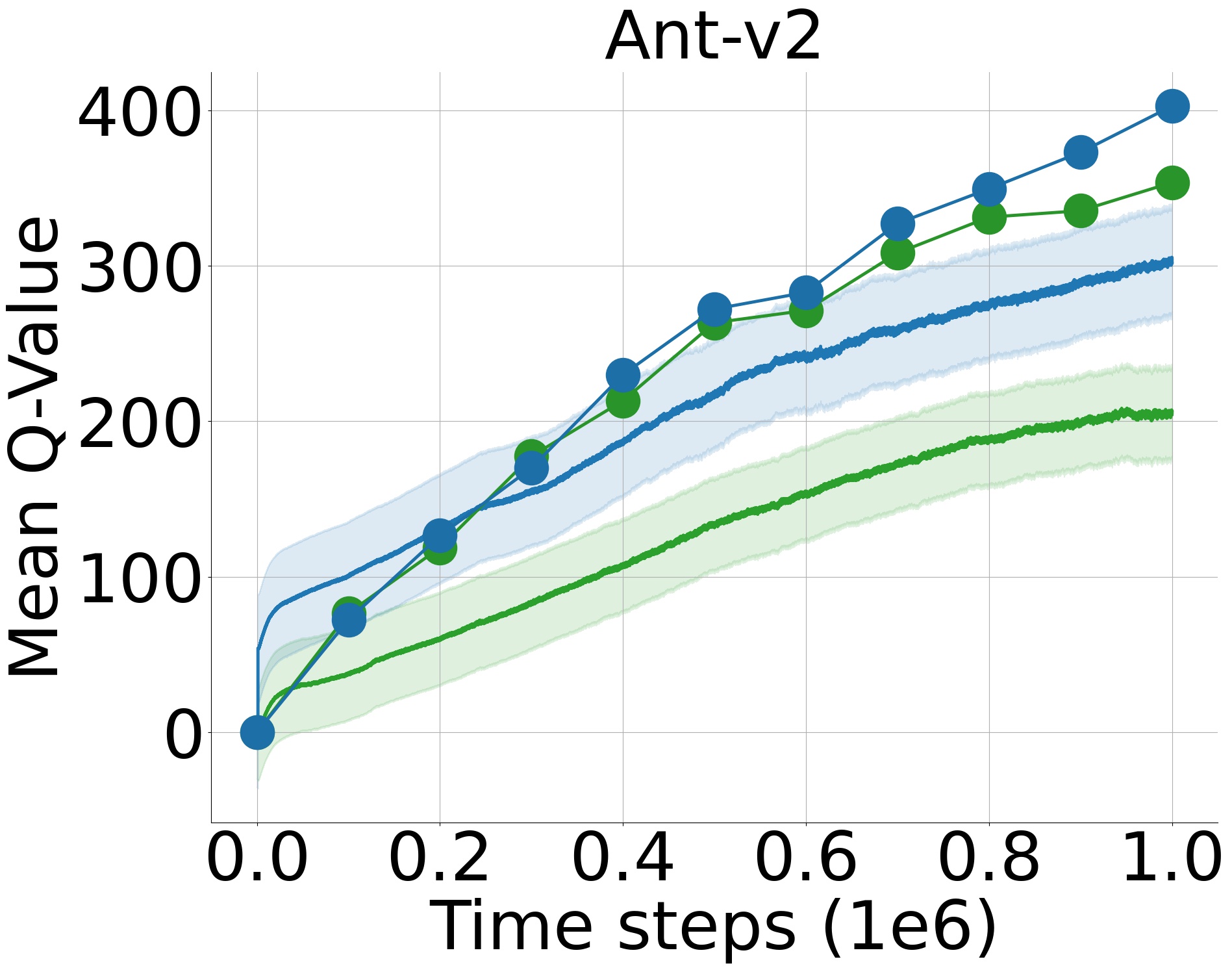}
		\includegraphics[width=1.01in, keepaspectratio]{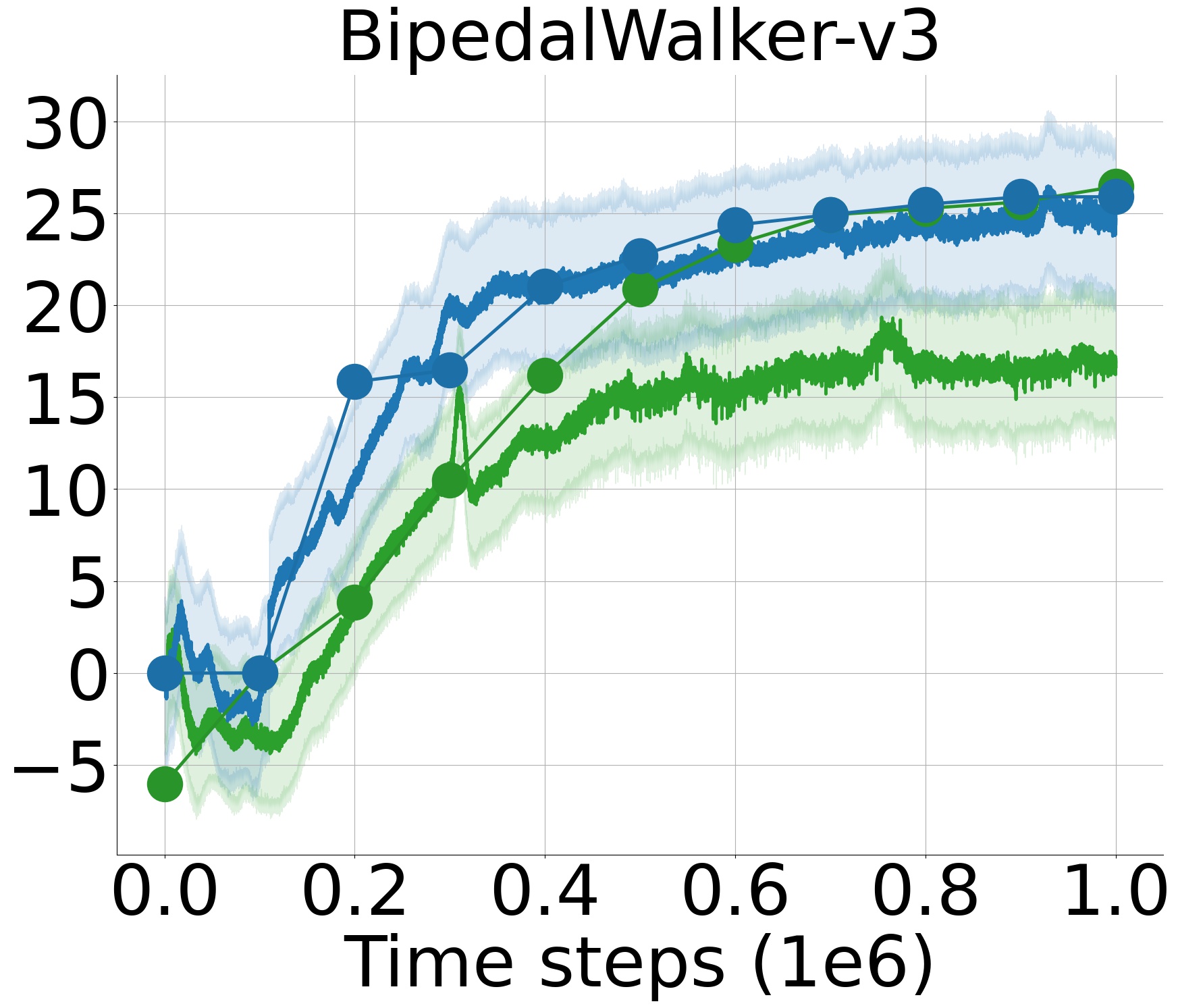}
		\includegraphics[width=1.01in, keepaspectratio]{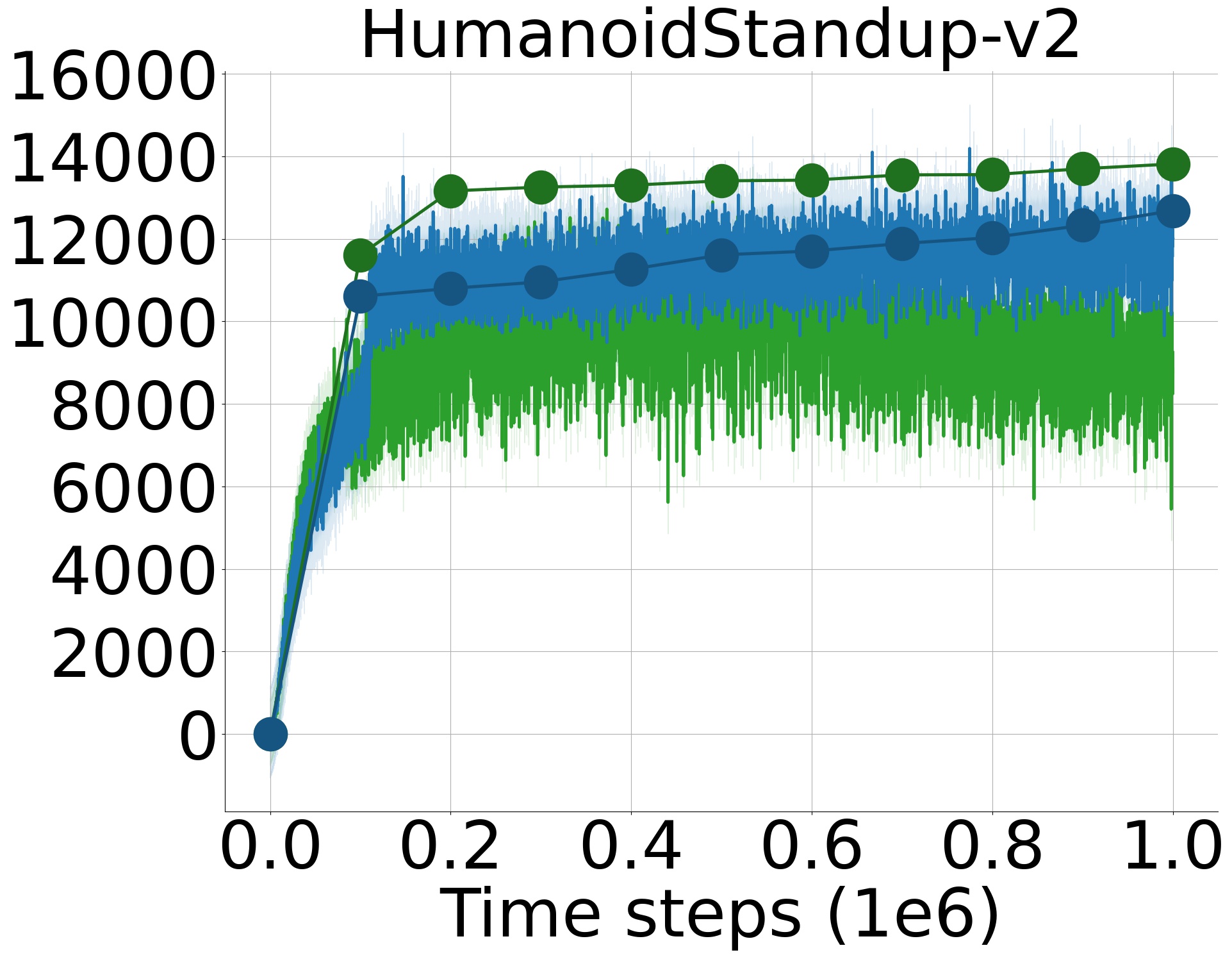}
			\includegraphics[width=1.01in, keepaspectratio]{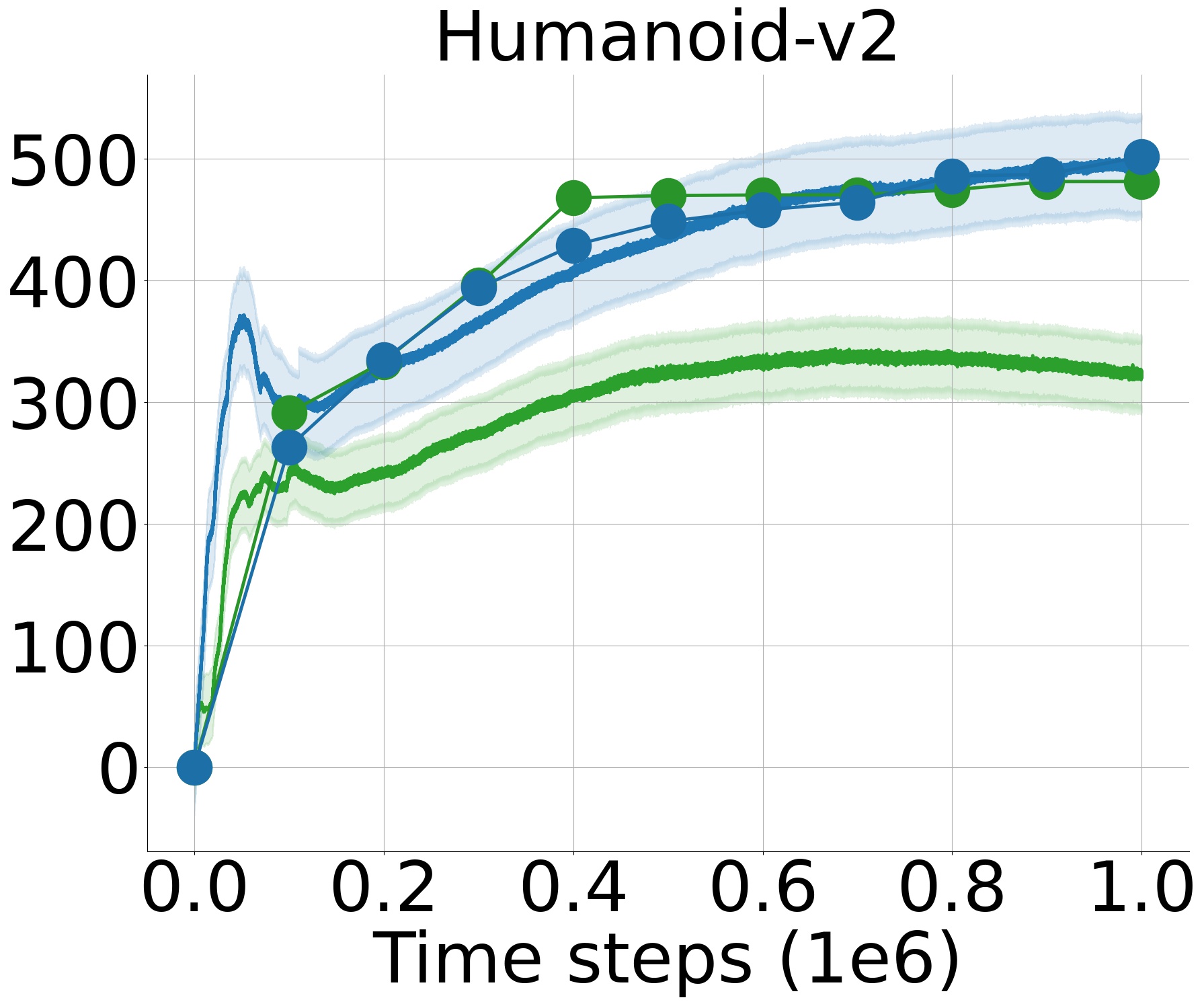}
		\includegraphics[width=1.01in, keepaspectratio]{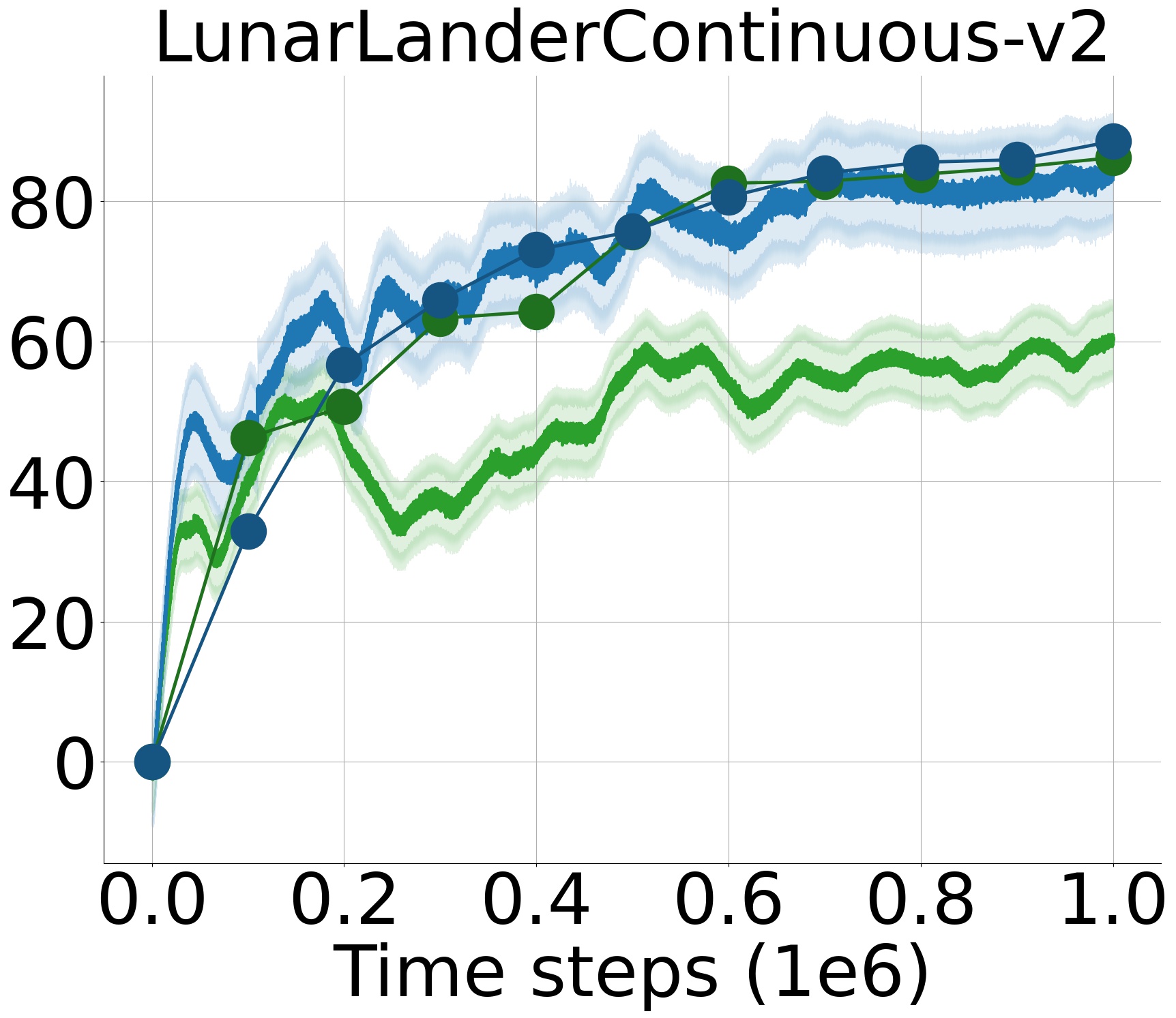}
		\includegraphics[width=1.01in, keepaspectratio]{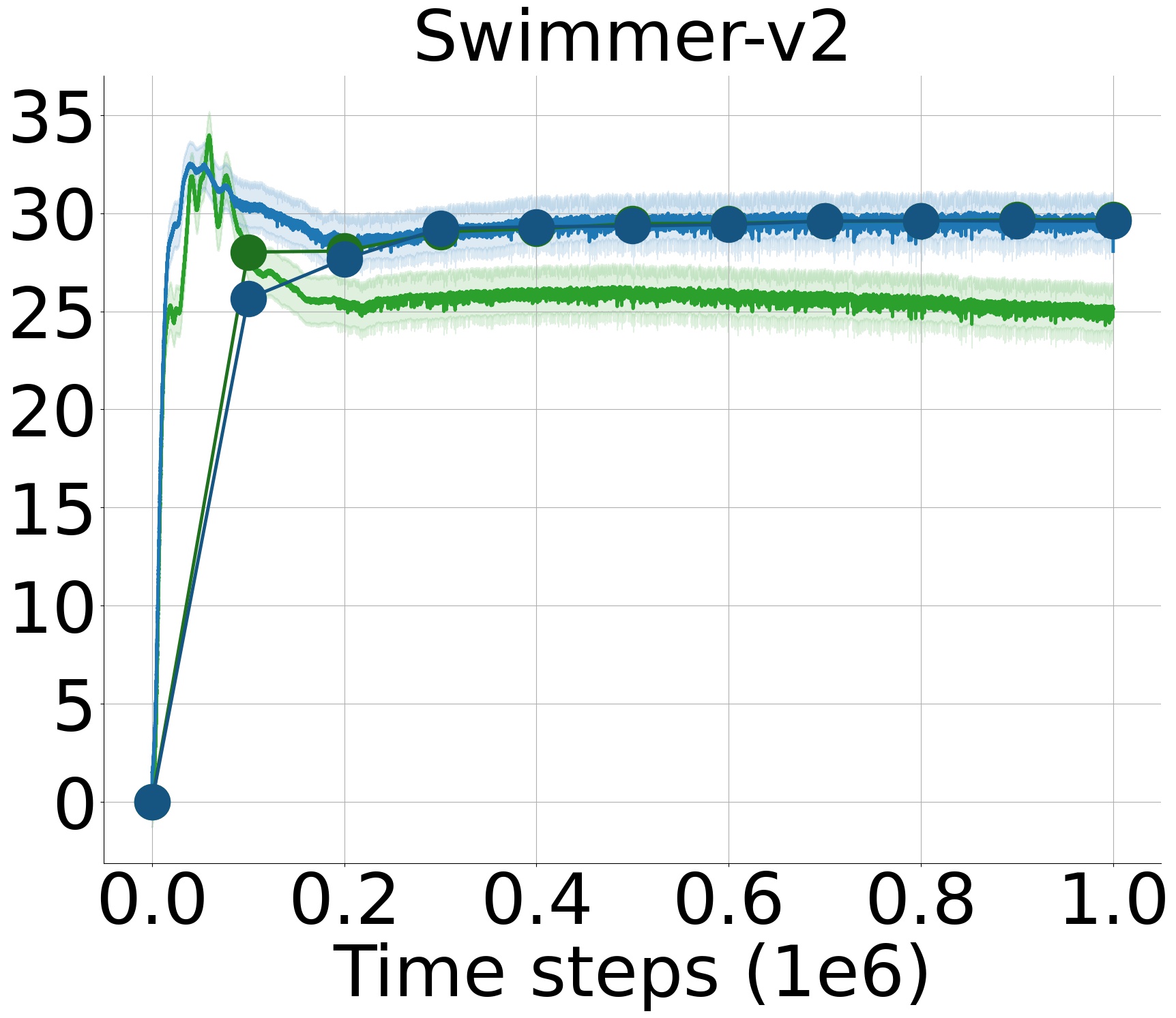}
	} 
	\caption{Measuring estimation bias of fine-tuned TD3 versus SWTD3 while learning on MuJoCo and Box2D environments over 1 million time steps. Estimated and true Q-values are computed through Monte Carlo simulation for 1000 samples.} 
	\label{td3_swtd_q_estimation}
\end{figure*}

We can show the existence of the underestimation bias in practice by comparing the true and estimated Q-values while an agent under the TD3 algorithm \cite{td3} is learning on a set of OpenAI Gym \cite{gym} continuous control tasks over a training duration of 1 million time steps. The simulation results are reported by Fig. \ref{td3_swtd_q_estimation}. We randomly select 1000 state-action pairs at every step and obtain the estimated Q-values by the first Q-network. The true Q-values are obtained at every 100,000 time steps by computing the discounted sum of rewards starting from a randomly sampled 1000 states following the current policy. The Monte-Carlo simulation \cite{monte_carlo} is used over the randomly selected states and state-action pairs to obtain the average true and estimated Q-values. 

\label{section:underestimation_problem}

From Fig. \ref{td3_swtd_q_estimation}, we observe an apparent underestimation bias throughout the learning phase such that the estimated Q-values are smaller than the true ones except for a small proportion of the initial time steps. The underestimation bias arises depending on the environment and either grows or settles to a fixed level. These simulation results verify our claims; the approximate critics overestimate the actual Q-values at the initial steps. However, when the agent starts exploring the environment and encounters varying rewards, the variance of the value estimates increases, and the underestimation bias starts growing. For BipedalWalker and LunarLanderContinuous, the underestimation bias becomes fixed after a duration. This is due to the span of the reward space. If the agent encounters a sufficiently large subspace at the beginning of the learning, the underestimation bias cannot become larger. However, suppose the agent does not receive a significantly large subspace. In that case, the underestimation bias keeps growing even with the delayed target and actor updates as in the rest of the environments. Although the continuous, multi-dimensional, and large state-action spaces contribute to the growth of error, the scale of the current RL benchmarks is still very small compared to the real-world tasks \cite{mach_1}. Hence, the underestimation bias will be more detrimental and inevitable when larger-scaled tasks are introduced. 

To overcome the shown underestimation bias, we first start by deriving the expected error induced by the update rule in TCD3 \cite{tcd3}, and reduce the number of Q-networks to two while obtaining the same expected error. Then, through an extensive analysis of the WD3 \cite{wd3} and TADD \cite{tadd} algorithms, we introduce our novel, hyper-parameter-free modification on the target Q-value update that can further reduce the underestimation bias while preventing the overestimation. 

\section{Algorithm}
\subsection{Methodology}
First, we consider the Q-network update rule in our previous work, the TCD3 algorithm \cite{tcd3}:
\begin{equation}
\label{eq:tcu}
    y = r + \gamma \mathrm{min}\left(\underset{i = 1, 2}{\mathrm{max}}(Q_{\theta'_{i}}(s', \pi_{\phi'}(s')), Q_{\theta'_{3}}(s', \pi_{\phi'}(s')\right),
\end{equation}
where we employed an additional third critic $Q_{\theta_{3}}$ with corresponding estimation error distribution $N_{3} \sim \mathcal{N}(\mu_{3}, \sigma_{3})$. As the first critic is used to optimize the policy and due to the randomness in transition sampling, the same probability distribution can represent the errors corresponding to the second and third critics, i.e., $N_{3} \sim \mathcal{N}(\mu_{2}, \sigma_{2})$. We previously showed that this update rule can upper- and lower-bound the Q-value estimates by taking the minimum of the maximum of the first two critics and the third critic. Now, let us derive the expected function approximation error induced by the Q-value target expressed by (\ref{eq:tcu}). First, expand $\mathrm{min}(\mathrm{max}(N_{1}, N_{2}), N_{3})$ in terms of the maximum of error Gaussian's:
\begin{align}
\label{tcu_expanded_expression}
    \begin{split}
    \mathrm{min}(\mathrm{max}(N_{1}, N_{2}), N_{3}) &= \frac{1}{2}\mathrm{max}(N_{1}, N_{2}) + \frac{1}{2}N_{3} \\ &-  \frac{1}{2}\vert\mathrm{max}(N_{1}, N_{2}) - N_{3}\vert.
    \end{split}
\end{align}
It is not trivial to compute the expectation of the latter term in the right-hand side of (\ref{tcu_expanded_expression}). However, we can rewrite (\ref{tcu_expanded_expression}) in terms of the maximum of three correlated Gaussian's and use the derivation for its expectation for equal means case from \cite{expect_of_moments_of_max}. For this purpose, let $N_{\mathrm{max}} = \mathrm{max}(\mathrm{max}(N_{1}, N_{2}), N_{3}) = \mathrm{max}(N_{1}, N_{2}, N_{3})$. Then, the expected value of (\ref{tcu_expanded_expression}) can be expressed as:
\begin{equation}
   \mathrm{min}(\mathrm{max}(N_{1}, N_{2}), N_{3}) + N_{\mathrm{max}} = \mathrm{max}(N_{1}, N_{2}) + N_{3},
\end{equation}
\begin{align}
   \begin{split}
        \mathbb{E}[\mathrm{min}(\mathrm{max}(N_{1}, N_{2}), N_{3})] &= \mathbb{E}[\mathrm{max}(N_{1}, N_{2})] + \mathbb{E}[N_{3}] \\ &- \mathbb{E}[N_{\mathrm{max}}].\label{tcu_raw_exp}   
   \end{split}
\end{align}
Under the assumption made in section \ref{section:underestimation_problem} that $\mu_{1} \approx \mu_{2} = \mu_{3}$, let us define $\mu \coloneqq \mu_{1} = \mu_{2} = \mu_{3}$. Now, we can directly import the special case for the expectation of maximum of correlated Gaussian's from \cite{expect_of_moments_of_max}. The equal means case states that if $N_{i} \sim \mathcal{N}(\mu, \sigma_{i})$, then the expected value of maximum of three Gaussian's can be expressed as:
\begin{equation}
\label{exp_max_of_three_gaussians}
    \mathbb{E}[\mathrm{max}(N_{1}, N_{2}, N_{3})] = \mu + \frac{1}{2\sqrt{2\pi}}(\theta_{1, 2} + \theta_{1, 3} + \theta_{2, 3}),
\end{equation}
where $\theta_{i, j} \coloneqq \sqrt{\sigma_{i}^{2} + \sigma_{j}^{2} - 2\rho\sigma_{i}\sigma_{j}}$. Due to the same experience replay \cite{exp_replay} used in updating the Q-networks and decoupled actor and the first critic, without loss of generality, we can further assume that $\theta \coloneqq \theta_{1, 2} = \theta_{1, 3} = \theta_{2, 3}$. Then, (\ref{exp_max_of_three_gaussians}) reduces to:
\begin{equation}
\label{last_exp_max_of_3_gaussians}
    \mathbb{E}[N_{\mathrm{max}}] = \mathbb{E}[\mathrm{max}(N_{1}, N_{2}, N_{3})] = \mu + \frac{3\theta}{2\sqrt{2\pi}}.
\end{equation}
Furthermore, using the exact distribution of $\mathbb{E}[\mathrm{max}(N_{1}, N_{2})]$ from \cite{nadaraj}, similar to (\ref{eq:nadaraj_min_exp}), we have:
\begin{align}
    \begin{split}
   \label{eq:exp_of_max_of_gaussians_raw}
    \mathbb{E}[\underset{i=1, 2}{\mathrm{max}}\{N_{i}\}] &= \mu_{2} + (\mu_{1} - \mu_{2})\Phi(\frac{\mu_{1} - \mu_{2}}{\theta}) \\ &+ \theta\psi(\frac{\mu_{1} - \mu_{2}}{\theta}). 
    \end{split}
\end{align}
Using the assumptions made, we can simplify (\ref{eq:exp_of_max_of_gaussians_raw}) into:
\begin{equation}
\label{last_exp_max_of_2_gaussians}
    \mathbb{E}[\mathrm{max}(N_{1}, N_{2})] = \mu + \frac{\theta}{\sqrt{2\pi}}.
\end{equation}
Inserting (\ref{last_exp_max_of_3_gaussians}), (\ref{last_exp_max_of_2_gaussians}) and $\mathbb{E}[N_{3}] = \mu$ into (\ref{tcu_raw_exp}), we derive:
    \begin{equation}
        \mathbb{E}[\mathrm{min}(\mathrm{max}(N_{1}, N_{2}), N_{3})] = \mu - \frac{\theta}{2\sqrt{2\pi}}.
    \end{equation}
Replacing $\mu$ with $\mu_{2}$, we can express expected function approximation error for $\mathrm{min}(\mathrm{max}(Q_{1}, Q_{2}), Q_{3})$ in terms of the expected error for the Clipped Double Q-learning \cite{td3} denoted by (\ref{eq:nadaraj_min_exp}) as:
\begin{equation}
\label{eq:expected_estimation_error_tcu}
    \mathbb{E}[\mathrm{min}(\mathrm{max}(N_{1}, N_{2}), N_{3})] = \frac{\mathbb{E}[\underset{i=1, 2}{\mathrm{min}}\{N_{i}\}] + \mu_{2}}{2}.
\end{equation}
This expected estimation bias is slightly less than the average of the underestimation in TD3 \cite{td3} and overestimation in the DDPG algorithm \cite{ddpg}. As the variance of the value estimates by two correlated critics are greater than the expected function approximation error, (\ref{eq:expected_estimation_error_tcu}) is still an underestimation. We can further reduce this underestimation by replacing $\mu_{2}$ with $\mu_{1}$ in (\ref{eq:expected_estimation_error_tcu}) as $\mu_{1} \geq \mu_{2} \geq 0$:
\begin{equation}
\label{eq:stochastic_weighted_twin_critic_update}
\text{\small$
    y = r + \frac{\gamma}{2} \left(\underset{i = 1, 2}{\mathrm{min}}(Q_{\theta'_{i}}(s', \pi_{\phi'}(s')) + Q_{\theta'_{1}}(s', \pi_{\phi'}(s'))\right).$}
\end{equation}
Observe that the expected value of (\ref{eq:tcu}) and (\ref{eq:stochastic_weighted_twin_critic_update}) are the same. We eliminate the computational burden introduced by the employment of the third Q-network while attaining the same expected error. Hence, the computational complexity is reduced by 33\%.

Now, let us show the expected error by the WD3 \cite{wd3} and TADD \cite{tadd} algorithms. Update rules in these methods were previously expressed in (\ref{eq:wd3_upt}) and (\ref{eq:tadd_upt}), respectively. Using the error Gaussian distributions in (\ref{eq:error_gaussian_def}) and expectation of minimum of two correlated Gaussians in (\ref{eq:exp_min}), expected error of WD3 \cite{wd3} is expressed as:
\begin{align}
\begin{split}
    \mathbb{E}[\beta \underset{i = 1, 2}{\mathrm{min}}N_{i} + \frac{1 - \beta}{2}\sum_{i = 1}^{2}\mu_{i}] &= \beta\mu_{1} - \beta\frac{\theta}{\sqrt{2\pi}} \\ &+ \frac{1 - \beta}{2}\sum_{i = 1}^{2}N_{i} \\
    &\approx \beta\mu - \beta\frac{\theta}{\sqrt{2\pi}} \\ &+ (1 - \beta)\mu, \\
    &= \mu - \beta\frac{\theta}{\sqrt{2\pi}}\label{eq:wd3_est_error}.
\end{split}
\end{align}
Note that (\ref{eq:wd3_est_error}) is satisfied as $\mu \approx \mu_{1} \approx \mu_{2} = \mu_{3}$. Similarly, TADD \cite{tadd} yields the following expected error:
\begin{align}
\begin{split}
    \mathbb{E}[\beta\underset{i = 1, 2}{\mathrm{min}}N_{i} + \frac{(1 - \beta)}{K}\sum_{k = 1}^{K}N_{3, k}] &= \beta\mu_{1} -  \beta\frac{\theta}{\sqrt{2\pi}} \\ &+ \frac{1 - \beta}{K}\sum_{k}\mu_{3, k}, \\
    &\approx \beta\mu - \beta\frac{\theta}{\sqrt{2\pi}} \\ &+ (1 - \beta)\mu, \\
    &= \mu - \beta\frac{\theta}{\sqrt{2\pi}}\label{eq:tadd_est_error},
\end{split}
\end{align}
where again, the latter equations are satisfied by $\mu \approx \mu_{1} \approx \mu_{2} = \mu_{3}$. Essentially, from (\ref{eq:wd3_est_error}) and (\ref{eq:tadd_est_error}), we observe that the estimation bias in the WD3 \cite{wd3} and TADD \cite{tadd} algorithms are the same. Moreover, by (\ref{eq:expected_estimation_error_tcu}), we can infer that the following equations hold:
\begin{align}
    \begin{split}
       \epsilon_{\mathrm{TCD3}} < \epsilon_{\mathrm{WD3}} = \epsilon_{\mathrm{TADD}} & \quad \mathrm{if} \quad \beta, < 0.5, \\ 
       \epsilon_{\mathrm{TCD3}} \geq \epsilon_{\mathrm{WD3}} = \epsilon_{\mathrm{TADD}} & \quad \mathrm{if} \quad \beta \geq 0.5,
    \end{split}
\end{align}
where $\epsilon$ denotes the estimation bias. We highlight our theoretical findings in the following Remarks. 
\begin{remark}
\label{rem:wd3_tadd_same_est_error}
   The Q-network update rule in the WD3 \cite{wd3} and TADD \cite{tadd} algorithms yield the same estimation bias.
\end{remark}
\begin{remark}
\label{rem:wd3_tadd_comp}
   If $\beta \geq 0.5$, the expected estimation bias induced by TCD3 \cite{tcd3} is larger than the bias induced by WD3 \cite{wd3} and TADD \cite{tadd}, and vice versa.
\end{remark}

Although the WD3 \cite{wd3} and TADD \cite{tadd} approaches violate Remark \ref{rem:zero_mean} and \ref{rem:iid}, utilizing a $\beta$ parameter enables the control of the underestimation bias. However, having a fixed $\beta$ is a task-specific greedy approach that cannot prevent the increasing underestimation bias as the variance of the reward signals increases throughout the learning. To overcome such an issue, we uniformly sample the $\beta$ parameter from an interval, the lower bound of which linearly decreases throughout the learning, consistent with the increasing variance of the reinforcement signals. To specify the upper and lower bounds for such sampling interval, we leverage the findings in our previous work \cite{tcd3}. In \cite{tcd3}, we showed that the estimation error by the Triplet Critic Update remains an underestimation, the absolute value of which is significantly smaller than that of Clipped Double Q-learning \cite{td3}. Although the estimation bias is not completely eliminated, the existing yet significantly decreased underestimation could dramatically improve the performance since underestimation is more preferable than overestimation \cite{td3}. As our previous work \cite{tcd3} corresponds to $\beta = 0.5$ in (\ref{eq:wd3_est_error}) and (\ref{eq:tadd_est_error}), we set the upper and lower bound of the interval to 0.5 in the beginning of the learning. Then, we linearly decrease the lower bound so that the contribution of an increasing variance of the rewards is also decreased throughout the learning. As we do not know the exact values of $\mu_{i}$ and $\theta$, yet we are sure that $\theta = 0$ yields overestimation, the final lower bound cannot be 0 but should be a small number, slightly larger than 0. For this, we set the final lower bound of the bias interval to a small number $\alpha = 0.05$. Formally, we obtain the $\beta$ parameter as:
\begin{gather}
    \beta^{\prime(0)} \leftarrow 0.5, \\
    \beta^{(0)} \sim [\beta^{\prime(0)}, \beta^{\prime(0)}], \\ 
    \beta^{(t)} \sim \mathrm{uniform}[\beta^{\prime(t)}, \beta^{\prime(0)}], \\ 
    \beta^{\prime(t + 1)} \leftarrow \beta^{\prime(0)} - \frac{\beta^{\prime(0)} - \alpha}{T} \times (t + 1),
\end{gather}
where $\beta^{(t)}$ is the sampled $\beta$ value at time step $t$, $\beta^{\prime(t)}$ is the lower bound of the sampling interval at time step $t$, and $T$ is the number of total training iterations. One concern with this update rule is that, as the exact estimation error cannot be known in theory, it may result in overestimation for some time steps. In addition, estimation error accumulates through subsequent updates in which Q-values are overestimated \cite{td3}. Nevertheless, the accumulated error will be clipped once a $\beta$ value that yields underestimation is sampled. Therefore, due to the randomness, the estimation error does not accumulate over a significant number of time steps throughout the learning, and the RL agents can tolerate such slightly overestimated Q-values \cite{tcd3}.

This forms our parameter-free update rule, Stochastic Weighted Twin Critic Update. As a result, our modification offers accurate Q-value estimates without introducing hyper-parameters and networks. We summarize our introduced approach in Algorithm \ref{alg:stochastic_weighted_twin_critic_update}, and the resulting algorithm built on the TD3 algorithm \cite{td3}, Stochastic Weighted Twin Delayed Deep Deterministic Policy Gradient (SWTD3) in Algorithm \ref{alg:swtd}.

\begin{remark}
   \label{rem:variance_does_not_affect}
   Due to the decreased lower bound of the $\beta$ sampling interval and hence the mean of the $\beta$ distribution, the introduced Q-network update rule is not affected as much as when $\beta$ is fixed. 
\end{remark}

\begin{remark}
\label{rem:swtd_overestimation}
    The estimation error induced by Stochastic Weighted Twin Critic Update may result in overestimation for some training iterations, especially in the later stages of learning, since the lower bound of the $\beta$ interval becomes very small. However, if a $\beta$ value corresponding to the underestimation is sampled, the overestimation will be clipped. Hence, estimation error does not accumulate over a significant number of time steps in the SWTD3 algorithm throughout learning due to its stochastic nature.
\end{remark}

\begin{algorithm*}[htbp]
    \caption{Stochastic Weighted Twin Critic Update (SWT)}
    \begin{algorithmic}[1]
        \STATE \textbf{Input:} $Q_{\theta_{1}'}, Q_{\theta_{2}'}$, $s'$, $\tilde{a}, \beta^{\prime(0)}, \beta^{\prime}, t, T$
        \STATE $\beta \sim \mathrm{uniform}[\beta^{\prime}, \beta^{\prime(0)}]$
        \STATE $y \leftarrow r + \gamma \left(\beta \underset{i = 1, 2}{\mathrm{min}}(Q_{\theta'_{i}}(s', \pi_{\phi'}(s')) + (1 - \beta)Q_{\theta'_{1}}(s', \pi_{\phi'}(s'))\right)$
        \STATE $\beta^{\prime} \leftarrow \beta^{\prime(0)} - \frac{\beta^{\prime(0)} - \alpha}{T} \times (t + 1)$ \\
        \RETURN $y, \beta^{\prime}$
    \end{algorithmic}
    \label{alg:stochastic_weighted_twin_critic_update}
\end{algorithm*}

\begin{algorithm*}[htbp]
    \caption{SWTD3}
    \begin{algorithmic}[1]
        \STATE Initialize critic networks $Q_{\theta_{1}}, Q_{\theta_{2}}$, and actor network $\pi_{\phi}$ with randomly initialized parameters $\theta_{1}, \theta_{2}, \phi$
        \STATE Initialize target networks $ \phi' \leftarrow \phi$, $\theta_{1}' \leftarrow \theta_{1}, \theta_{2}' \leftarrow \theta_{2}$
        \STATE Initialize replay buffer $\mathcal{B}$
        \STATE Initialize the lower bound of the $\beta$ sampling interval $\beta^{\prime(0)} \leftarrow 0.5$
        \FOR{$t = 1$ \textbf{to} $T$}
            \STATE Select action with exploration noise $a \sim \pi_{\phi}(s) + \mathcal{N}(0, \sigma)$, and observe reward $r$ and new state $s'$
            \STATE Store transition tuple $(s, a, r, s')$ in $\mathcal{B}$
            \STATE Sample mini-batch of $K$ transitions $(s, a, r, s')$ from $\mathcal{B}$
            \STATE $\tilde{a} \leftarrow \pi_{\phi'}(s') + \mathrm{clip}(\mathcal{N}(0, \sigma), -c, c)$
            \STATE $y, \beta^{\prime} \leftarrow \text{\textbf{SWT}}(Q_{\theta'_{1}}, Q_{\theta'_{2}}, s', \tilde{a}, \beta^{\prime(0)}, \beta^{\prime}, t, T)$ 
            \STATE Update critics $\theta_{i} \leftarrow \mathrm{argmin}_{\theta_{i}}\sum(y - Q_{\theta_{i}}(s, a))^{2} / K$
            \IF{$t$ mod $d$}
                \STATE Update $\phi$ by the deterministic policy gradient:
                \STATE $\nabla_{\phi}J(\phi)=\frac{1}{K}\sum\nabla_{a}Q_{\theta_{1}}(s, a)\vert_{a=\pi_{\phi(s)}}\nabla_{\phi}\pi_{\phi}(s)$
                \STATE Update target networks:
                \STATE $\theta_{i}' \leftarrow \tau\theta_{i} + (1 - \tau)\theta_{i}'$
                \STATE $\phi' \leftarrow \tau\phi + (1 - \tau)\phi'$
            \ENDIF
        \ENDFOR
    \end{algorithmic}
    \label{alg:swtd}
\end{algorithm*}

\subsection{Algorithmic and Complexity Comparison with the Existing Strategies}
We investigate how our method differs from the previously examined approaches to the underestimation bias. First, we derive our method by assuming positively biased Q-value estimators and dependence of the approximate critics, which are mandatory and realistic in practice. These requirements were previously summarized in Remark \ref{rem:zero_mean} and \ref{rem:iid}, respectively. Second, our method does not introduce any hyper-parameter to be tuned in contrast to the WD3 \cite{wd3} and TADD \cite{tadd} algorithms that require the $\beta$ parameter to be tuned, which controls the underestimation.

As we explained previously, the TD3 \cite{td3} and WD3 \cite{wd3} algorithms maintain two critics while TADD \cite{tadd} trains three critics. Although the Q-network objective computation requires the estimation of target Q-networks, the behavioral Q-networks must be maintained as the soft or hard update is used to update the corresponding target networks. Moreover, the TADD algorithm \cite{tadd} uses estimations of $K$ target Q-networks in constructing the Q-network objective. Nevertheless, the time complexity of backpropagation through a network either matches or is larger than the forward propagation. Hence, we consider the time complexity as the only number of backpropagated Q-networks. Therefore, our method, TD3 \cite{td3} and WD3 \cite{wd3} match in terms of the run time and are bounded by the time complexity of TADD \cite{tadd}. The following Remarks are made to conclude this comparison. s

\begin{remark}
   Our method introduces an analytical solution to the underestimation bias for deterministic policy gradients by considering biased Q-value estimators and dependence of the Q-networks in Clipped Double Q-learning \cite{td3}, contrasting with the WD3 \cite{wd3} and TADD \cite{tadd} studies.
\end{remark}

\begin{remark}
   Our method does not introduce any hyper-parameter to be optimized, in contrast to WD3 \cite{wd3} and TADD \cite{tadd}, in which the underestimation control parameter $\beta$ requires to be tuned for each continuous task. 
\end{remark}

\begin{remark}
   Time complexity of the TD3 \cite{td3}, WD3 \cite{wd3} and SWTD3 algorithms match and are bounded by the time complexity of the TADD algorithm \cite{tadd}. 
\end{remark}

\section{Experiments}
We evaluate the performance of our estimation bias correction approach by first demonstrating the estimated and actual Q-values of SWTD3 versus TD3 \cite{td3}, WD3 \cite{wd3} and TADD \cite{tadd}. Then, we evaluate the learning performances of RL agents under the SWTD3, TD3 \cite{td3}, WD3 \cite{wd3} and TADD \cite{tadd} algorithms in MuJoCo \cite{mujoco} and Box2D \cite{box2d} continuous control tasks interfaced by OpenAI Gym\footnote{\url{https://gym.openai.com/}} \cite{gym}. We also consider our previous work, TCD3 \cite{tcd3}, in our comparative evaluations for discussion. For reproducibility and a fair evaluation procedure, we directly follow the same set of tasks from MuJoCo \cite{mujoco} and Box2D \cite{box2d} with no modifications on the environment dynamics. 

\subsection{Implementation Details and Experimental Setup}
To implement the TD3 algorithm \cite{td3}, we use the author's GitHub repository\footnote{\url{https://github.com/sfujim/TD3}\label{repo:td3}}. The implementation of TD3 \cite{td3} is the fine-tuned version of the algorithm. This version of TD3 \cite{td3} differs from the one introduced in \cite{td3} such that the number of hidden units in all networks is reduced to 256, the batch size is increased from 100 to 256, learning rates for the behavioral actor and critic Adam optimizers \cite{adam} are decreased from $10^{-3}$ to $3\times10^{-4}$, and 25000 time steps of pure exploratory policy is employed in all environments. Furthermore, we built our modification on the TD3 \cite{td3} implementation such that the target Q-value computation is replaced by Algorithm \ref{alg:stochastic_weighted_twin_critic_update}. To ensure stability over updates and for a consistency with our theoretical approach, the actor in SWTD3 is always optimized with respect to the first critic, as in the TD3 \cite{td3} and TCD3 \cite{tcd3} algorithms. 

\begin{figure*}[ht]
	\centering
	\begin{equation*}
	    \text{{\blue} SWTD3} \quad \text{\textcolor{custom_blue}{$\boldsymbol{\bullet}$} True SWTD3} \qquad \text{{\red} WD3} \quad \text{\textcolor{custom_red}{$\boldsymbol{\bullet}$} True WD3}
	\end{equation*}
	\subfigure{
		\includegraphics[width=1.01in, keepaspectratio]{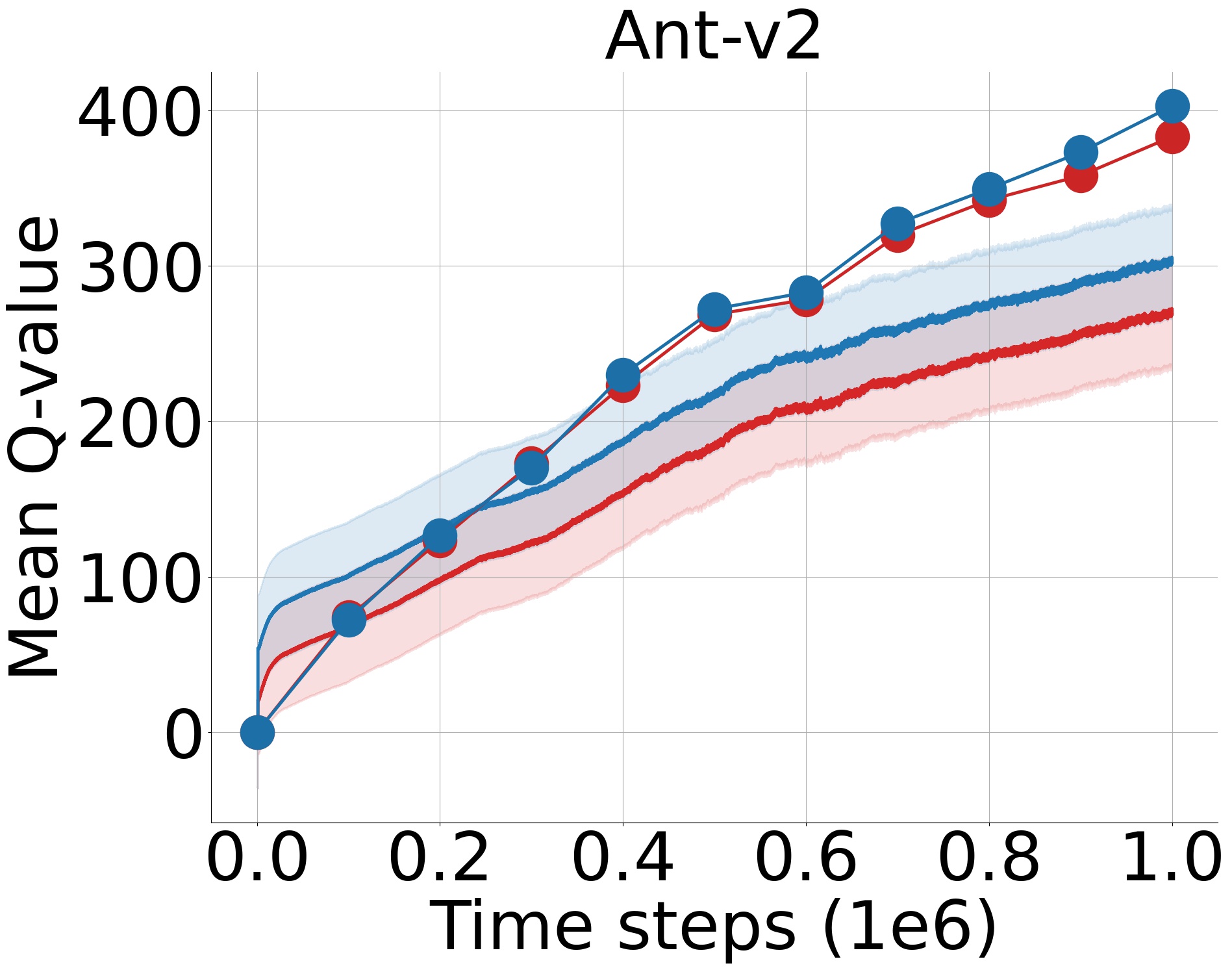}
		\includegraphics[width=1.01in, keepaspectratio]{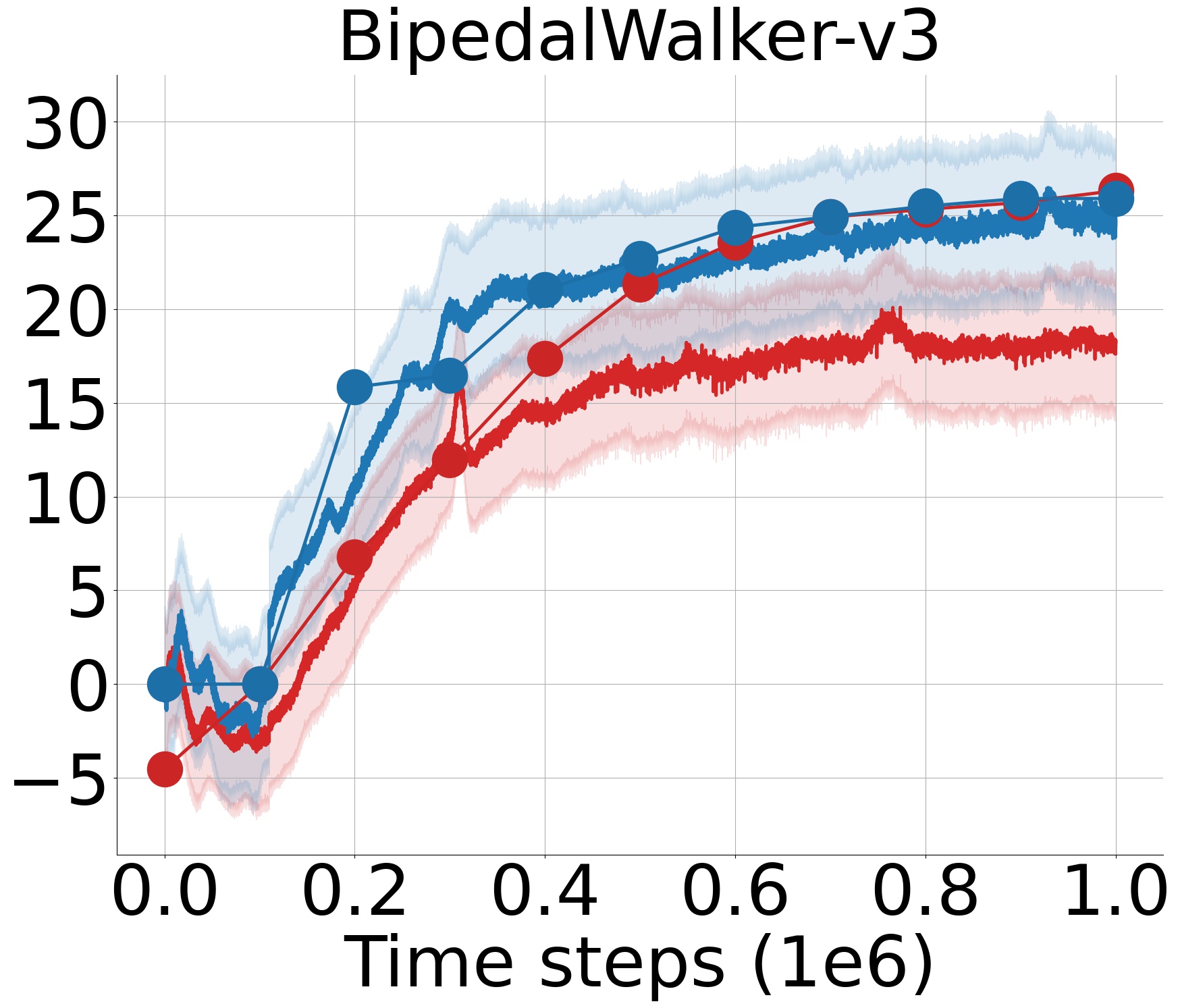}
		\includegraphics[width=1.01in, keepaspectratio]{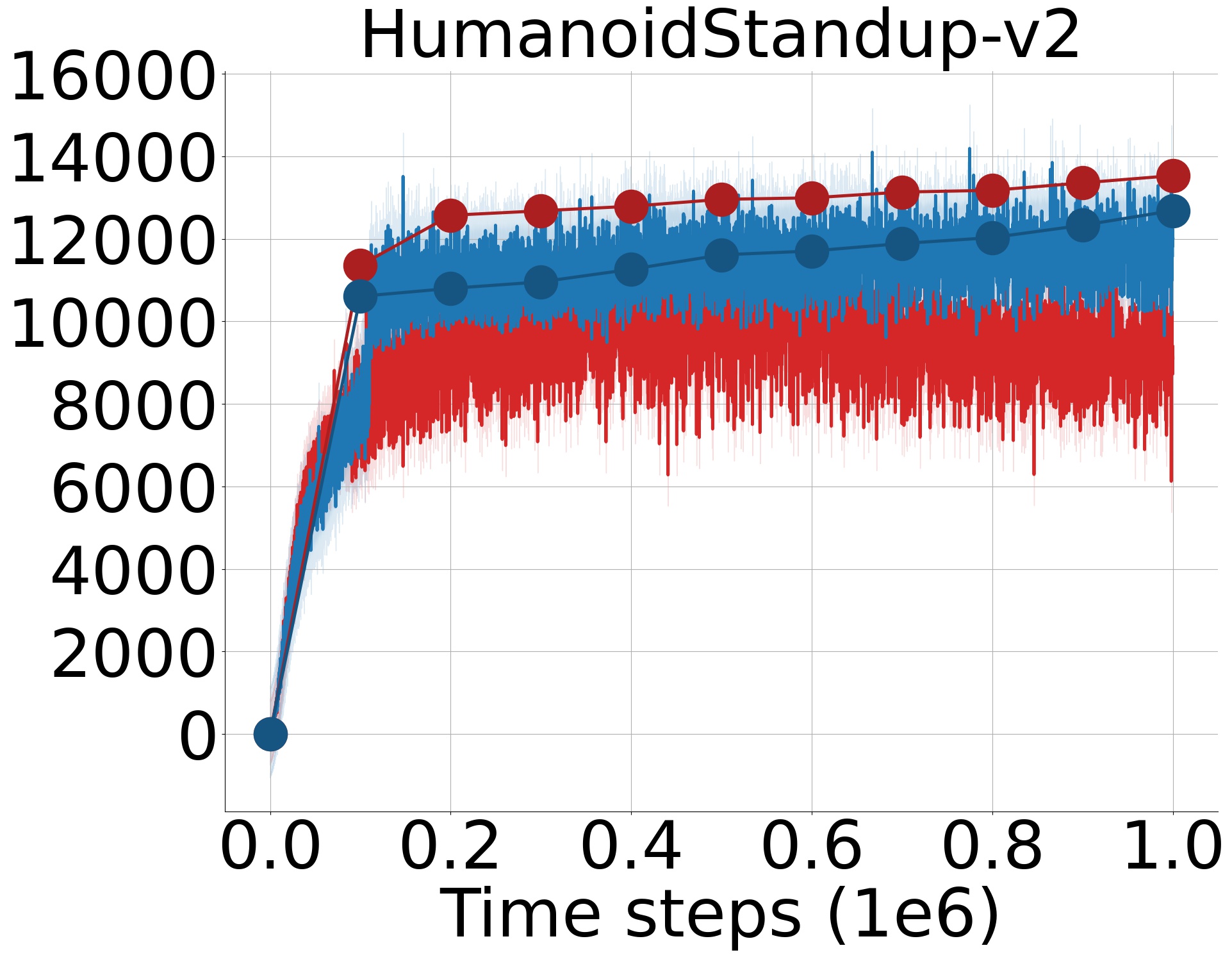}
		\includegraphics[width=1.01in, keepaspectratio]{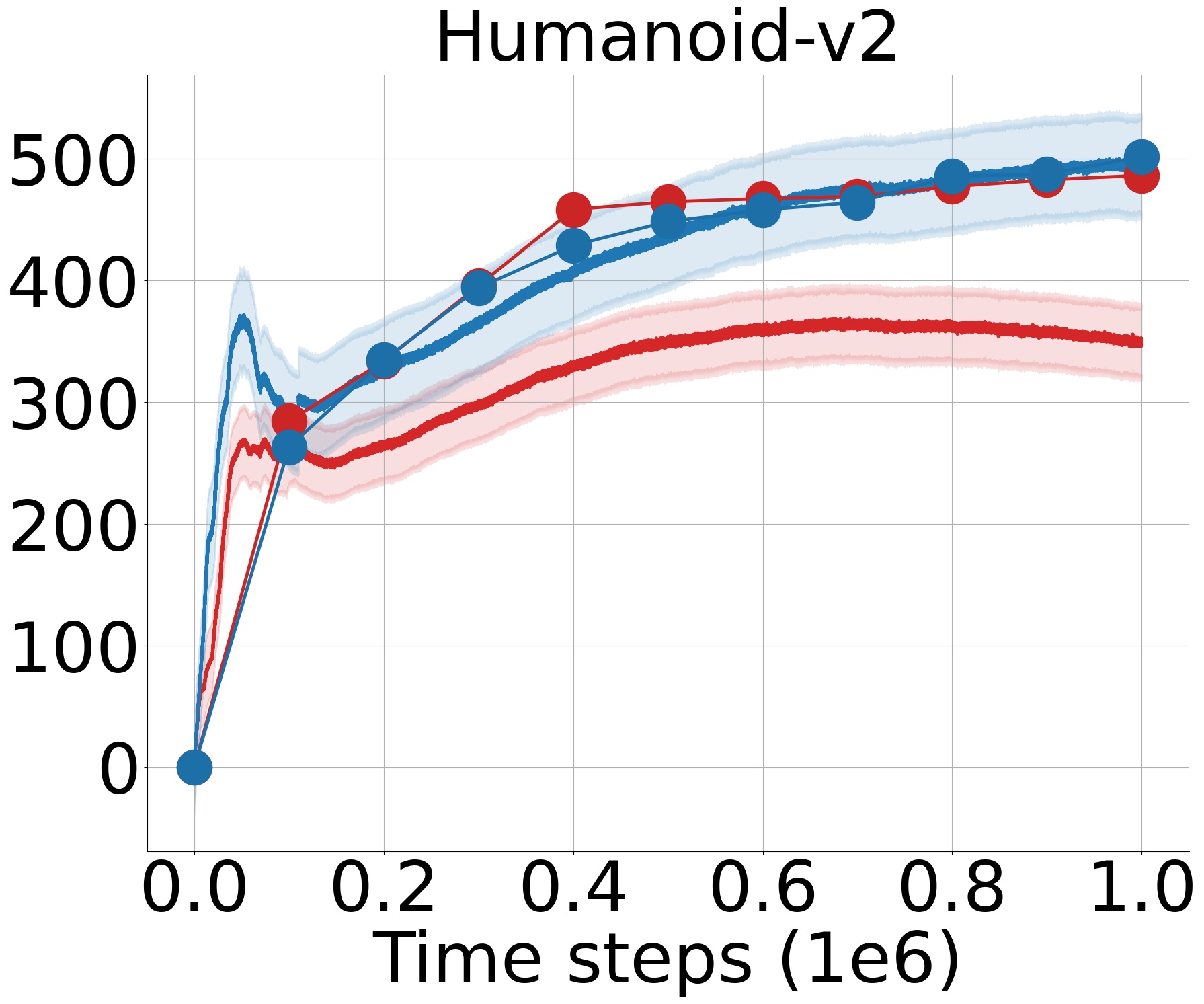}
		\includegraphics[width=1.01in, keepaspectratio]{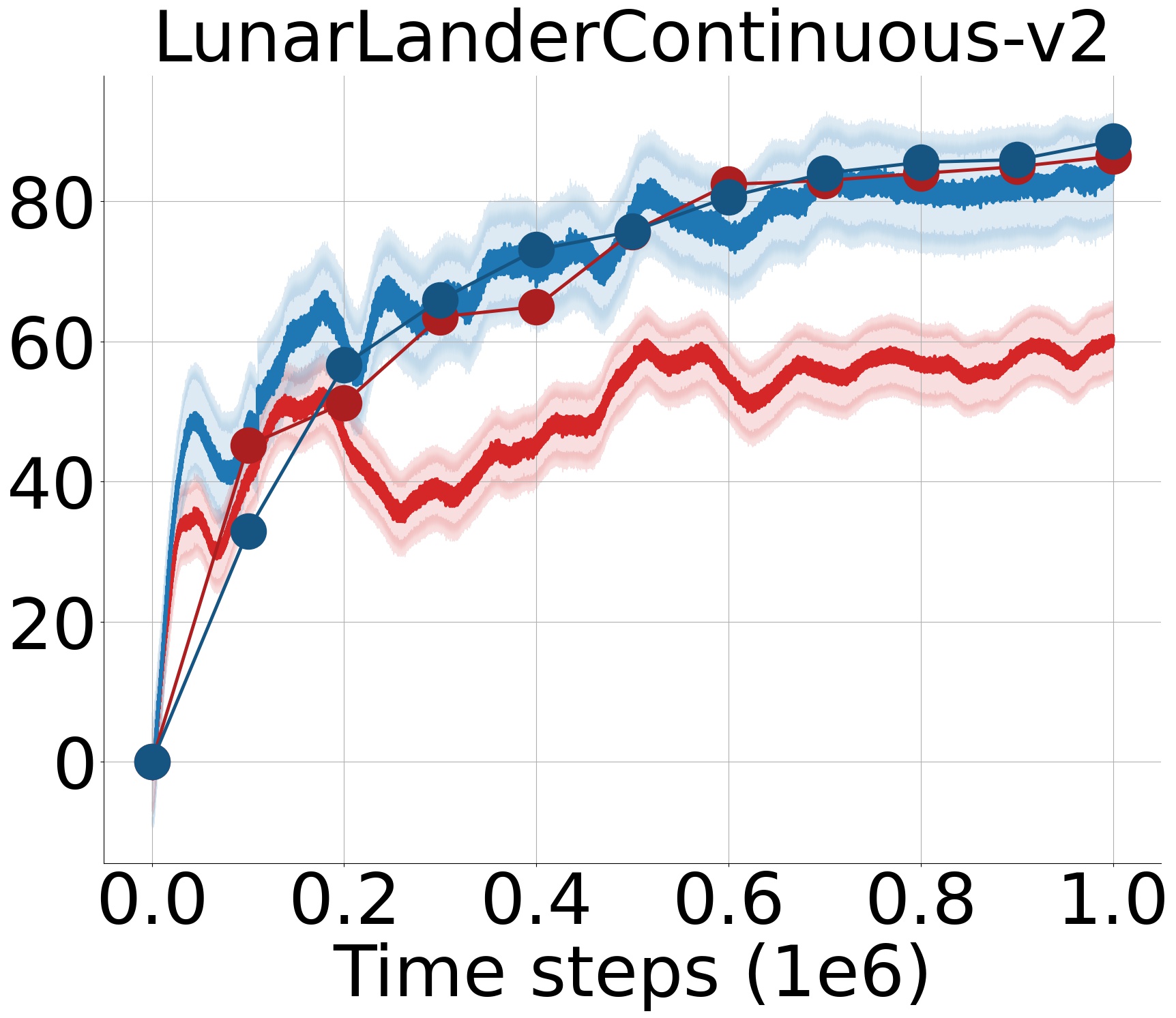}
		\includegraphics[width=1.01in, keepaspectratio]{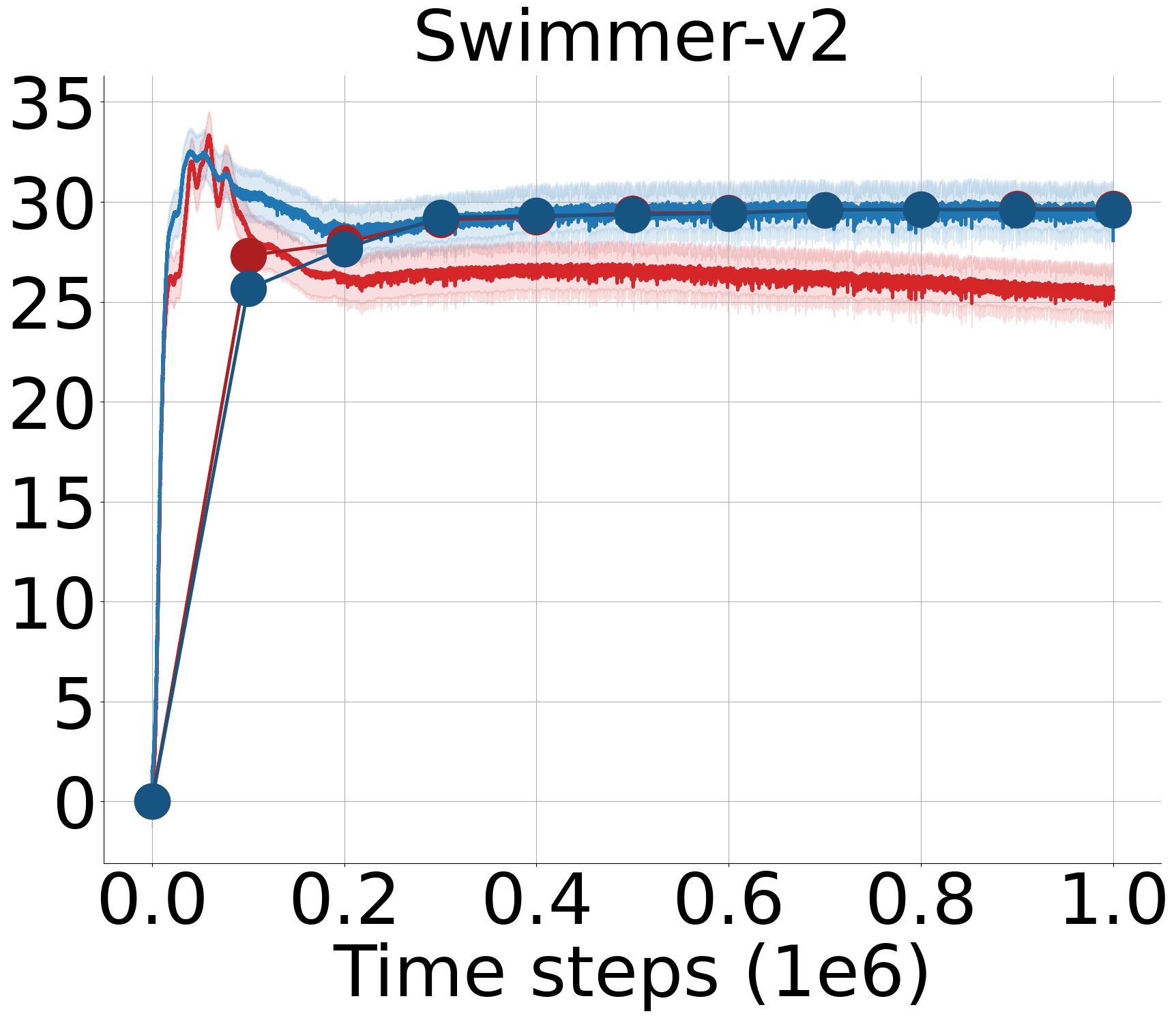}
	} 
	\caption{Measuring estimation bias produced WD3 versus SWTD3 while learning on MuJoCo and Box2D environments over 1 million time steps. Estimated and true Q-values are computed through Monte Carlo simulation for 1000 samples.}
	\label{wd3_swtd_q_estimation}
\end{figure*}

\begin{figure*}[ht]
	\centering
	\begin{equation*}
	    \text{{\blue} SWTD3} \quad \text{\textcolor{custom_blue}{$\boldsymbol{\bullet}$} True SWTD3} \qquad \text{{\orange} TADD} \quad \text{\textcolor{custom_orange}{$\boldsymbol{\bullet}$} True TADD}
	\end{equation*}
	\subfigure{
		\includegraphics[width=1.01in, keepaspectratio]{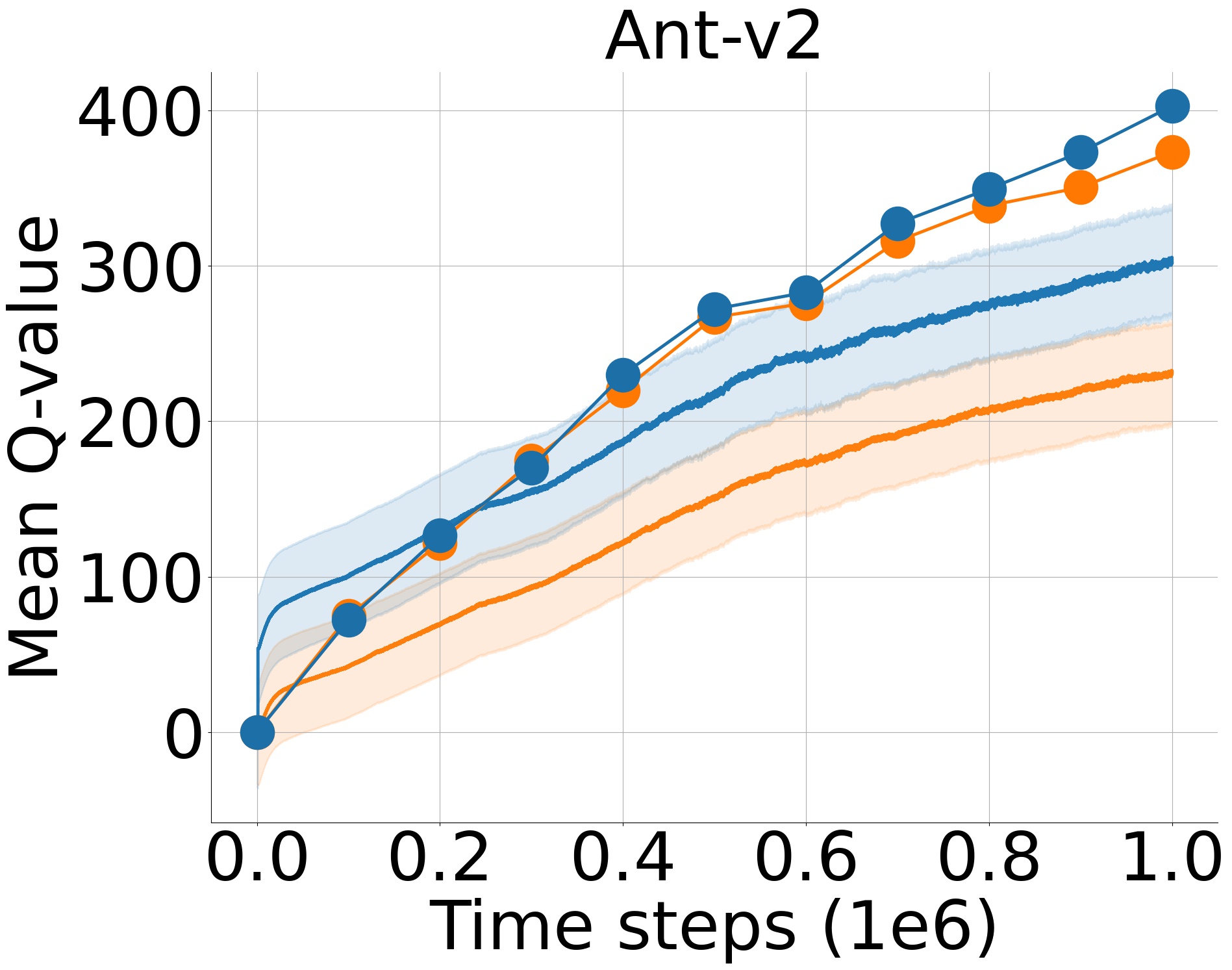}
		\includegraphics[width=1.01in, keepaspectratio]{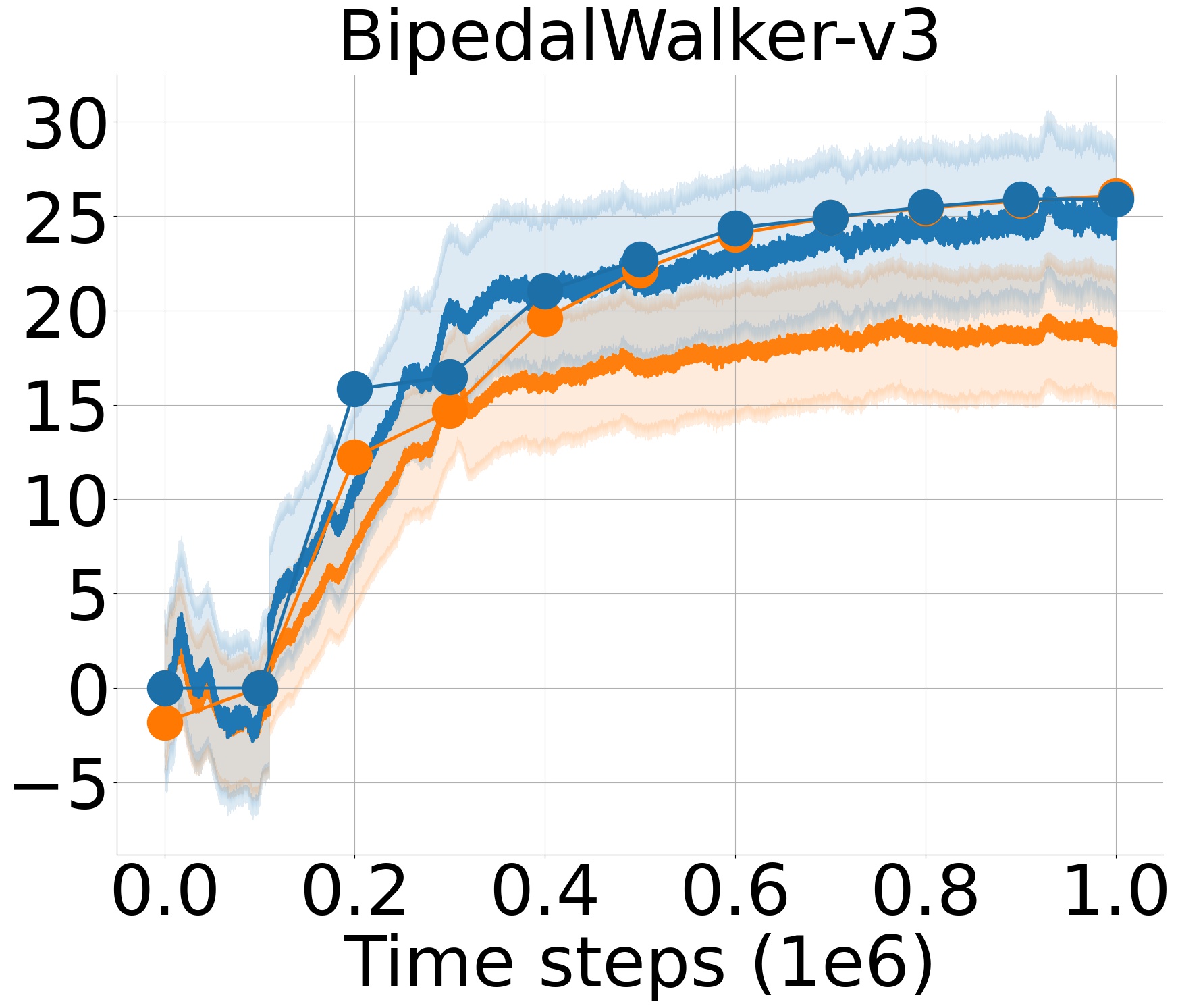}
		\includegraphics[width=1.01in, keepaspectratio]{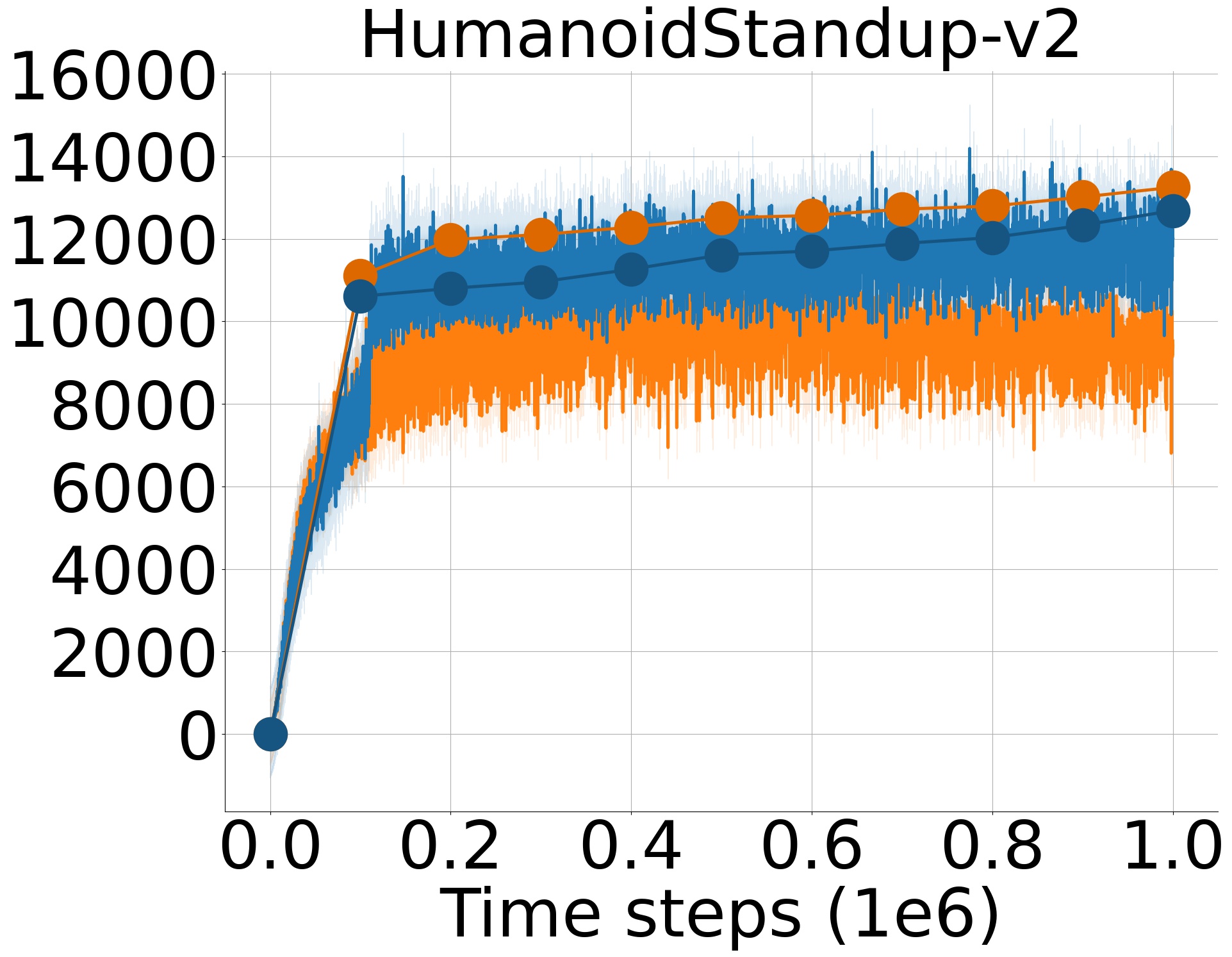}
		\includegraphics[width=1.01in, keepaspectratio]{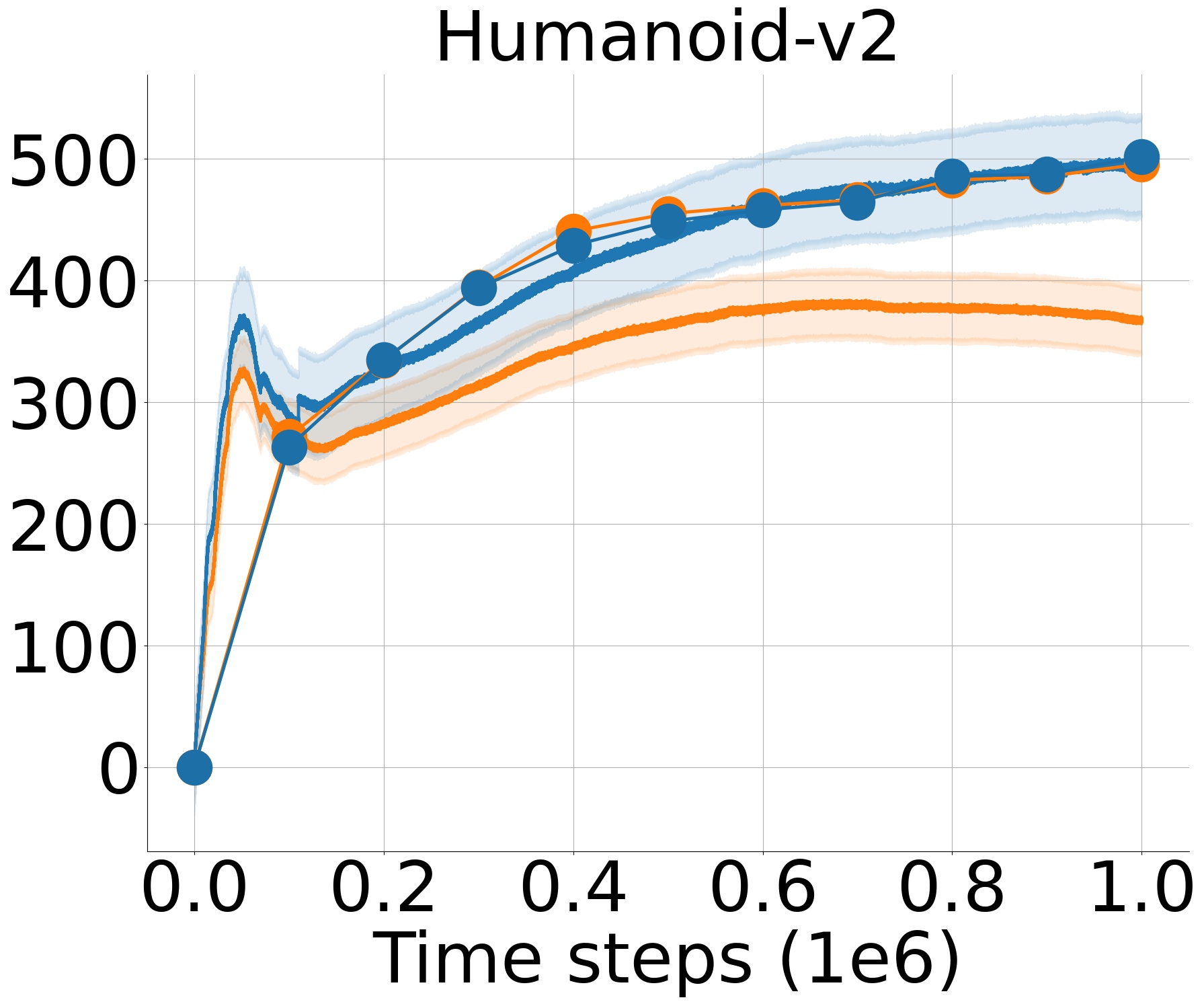}
		\includegraphics[width=1.01in, keepaspectratio]{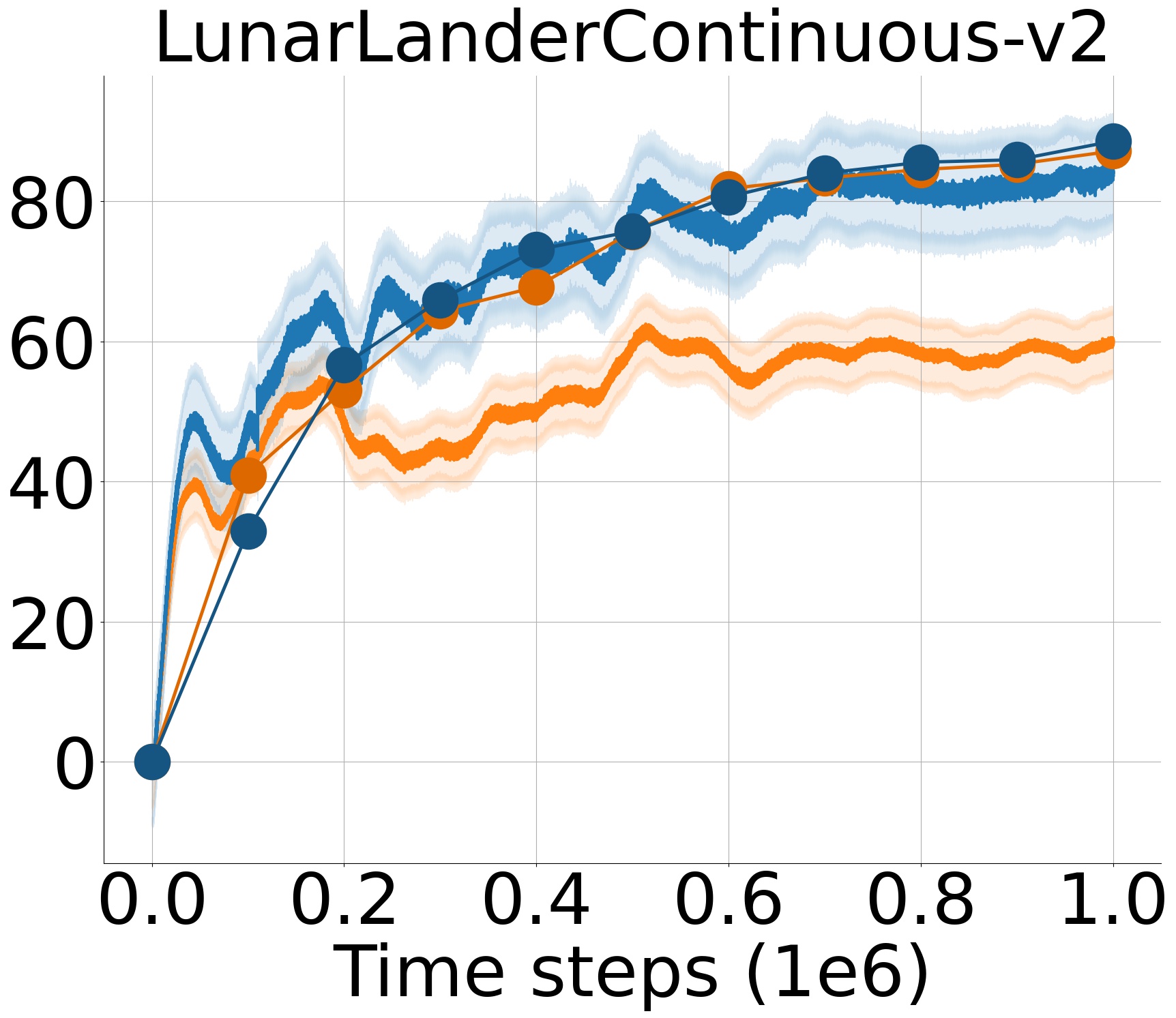}
		\includegraphics[width=1.01in, keepaspectratio]{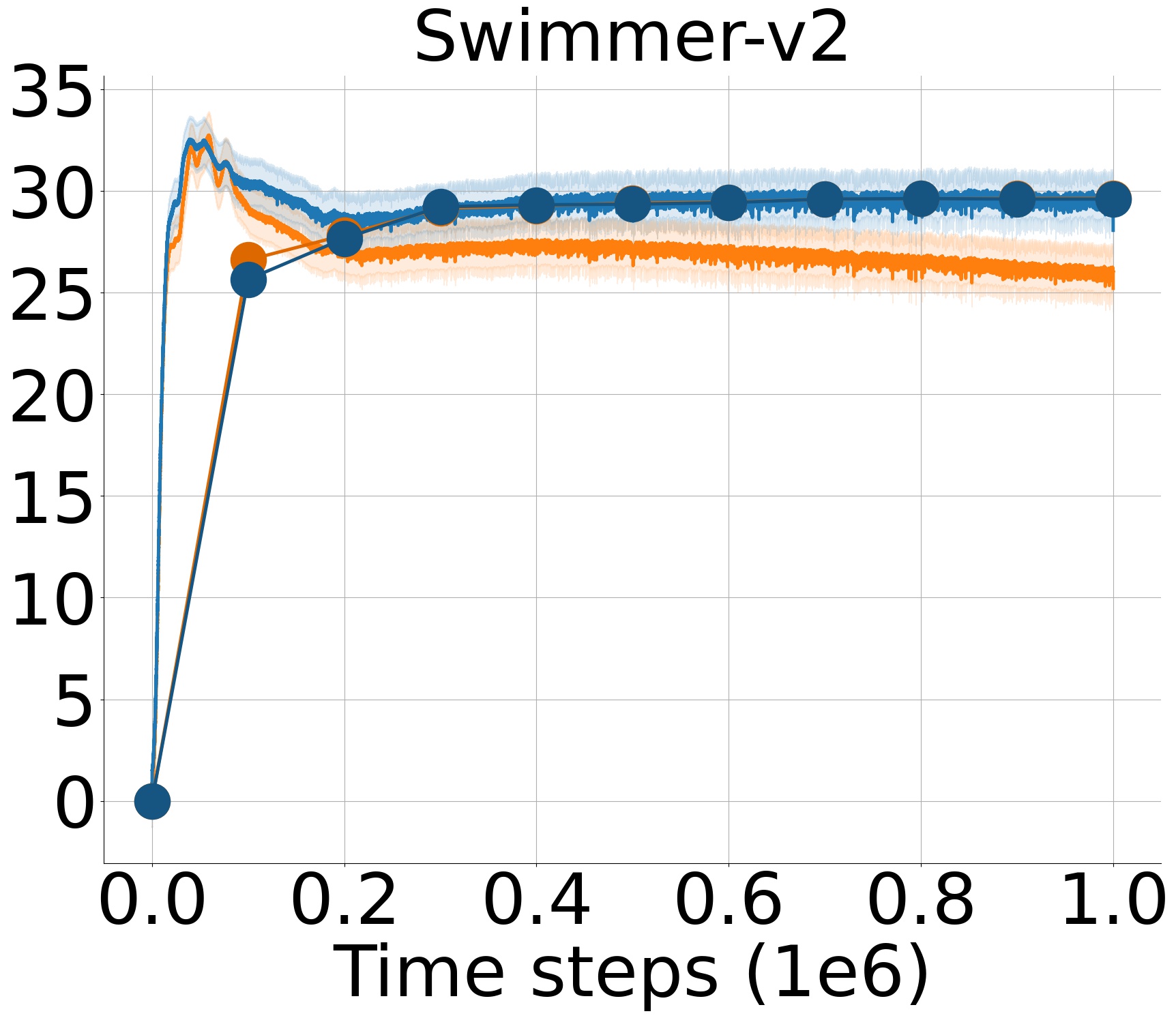}
	}
	\caption{Measuring estimation bias produced TADD versus SWTD3 while learning on MuJoCo and Box2D environments over 1 million time steps. Estimated and true Q-values are computed through Monte Carlo simulation for 1000 samples.}
	\label{tadd_swtd_q_estimation}
\end{figure*}

To implement the baseline algorithms, WD3 \cite{wd3} and TADD \cite{tadd}, we use the TD3 algorithm's repository. We follow the same parameter, network, and Q-value update structures in \cite{wd3} and \cite{tadd} such that we replace the target Q-value computation and initialize an additional Q-network if required. For the pre-defined weight parameter $\beta$, we use the values for the environments presented in the respective papers. We manually fine-tune the $\beta$ value over a training duration of 1 million time steps for ten random seeds for the rest of the environments. The values with the highest average of the last ten evaluation return over ten random seeds are chosen to train WD3 \cite{wd3} and TADD \cite{tadd} algorithms. Table \ref{wd3_beta_values_table} presents the used environment-specific weight parameter $\beta$ values for the WD3 \cite{wd3} and TADD \cite{tadd} algorithms. Values that we fine-tune and presented in \cite{wd3} and \cite{tadd} are marked.

\begin{table}[!hbt]
\begin{center}
\caption{WD3 and TADD environment specific weight values.}
\label{wd3_beta_values_table}
\begin{threeparttable}
    \begin{tabular}{@{} lccc @{}}
        \toprule
        \textbf{Environment} & \textbf{WD3} & \textbf{TADD} \\
        \midrule
        Ant-v2 & 0.75\tnote{a} & 0.95\tnote{a} \\ 
        BipedalWalker-v3 & 0.5\tnote{b} & 0.5\tnote{b} \\ 
        HalfCheetah-v2 & 0.45\tnote{a} & 0.95\tnote{a} \\ 
        Hopper-v2 & 0.50\tnote{a} & 0.95\tnote{a} \\ 
        HumanoidStandup-v2 & 0.30\tnote{b} & 0.30\tnote{b}\\ 
        Humanoid-v2 & 0.30\tnote{b} & 0.30\tnote{b}\\ 
        InvertedDoublePendulum-v2 & 0.75\tnote{a} & 0.95\tnote{a}\\ 
        InvertedPendulum-v2 & 0.75\tnote{a} & 0.95\tnote{a}\\ 
        LunarLanderContinuous-v2 & 0.45\tnote{b} & 0.45\tnote{b}\\ 
        Reacher-v2 & 0.15\tnote{a} & 0.95\tnote{a} \\ 
        Swimmer-v2 & 0.45\tnote{b} & 0.20\tnote{a} \\ 
        Walker2d-v2 & 0.45\tnote{a} & 0.95\tnote{a}\\   
        \bottomrule
    \end{tabular}
    \begin{tablenotes}    
        \item[a]{As given in the paper}
        \item[b]{Fine-tuned}
    \end{tablenotes}
\end{threeparttable}
\end{center}
\end{table}

Each task in the Q-value comparisons is run for 1 million time steps, and curves are derived through the same procedure explained in section \ref{section:underestimation_problem}. We perform evaluations on every task by running the algorithms over 1 million time steps and evaluating the agent's performance in a distinct evaluation environment without exploration noise and learning at every 1000 time steps. Each evaluation report is an average of ten episode rewards. The results are reported over ten random seeds of the Gym \cite{gym} simulator, network initialization, and code dependencies.

\begin{figure*}[!htbp]
	\centering
	\begin{equation*}
	    \text{{\blue} SWTD3} \quad \text{{\orange} TADD} \quad \text{{\red} WD3} \quad \text{{\brown} TCD3} \quad \text{{\green} TD3}
	\end{equation*}
	\subfigure{
		\includegraphics[width=1.6in, keepaspectratio]{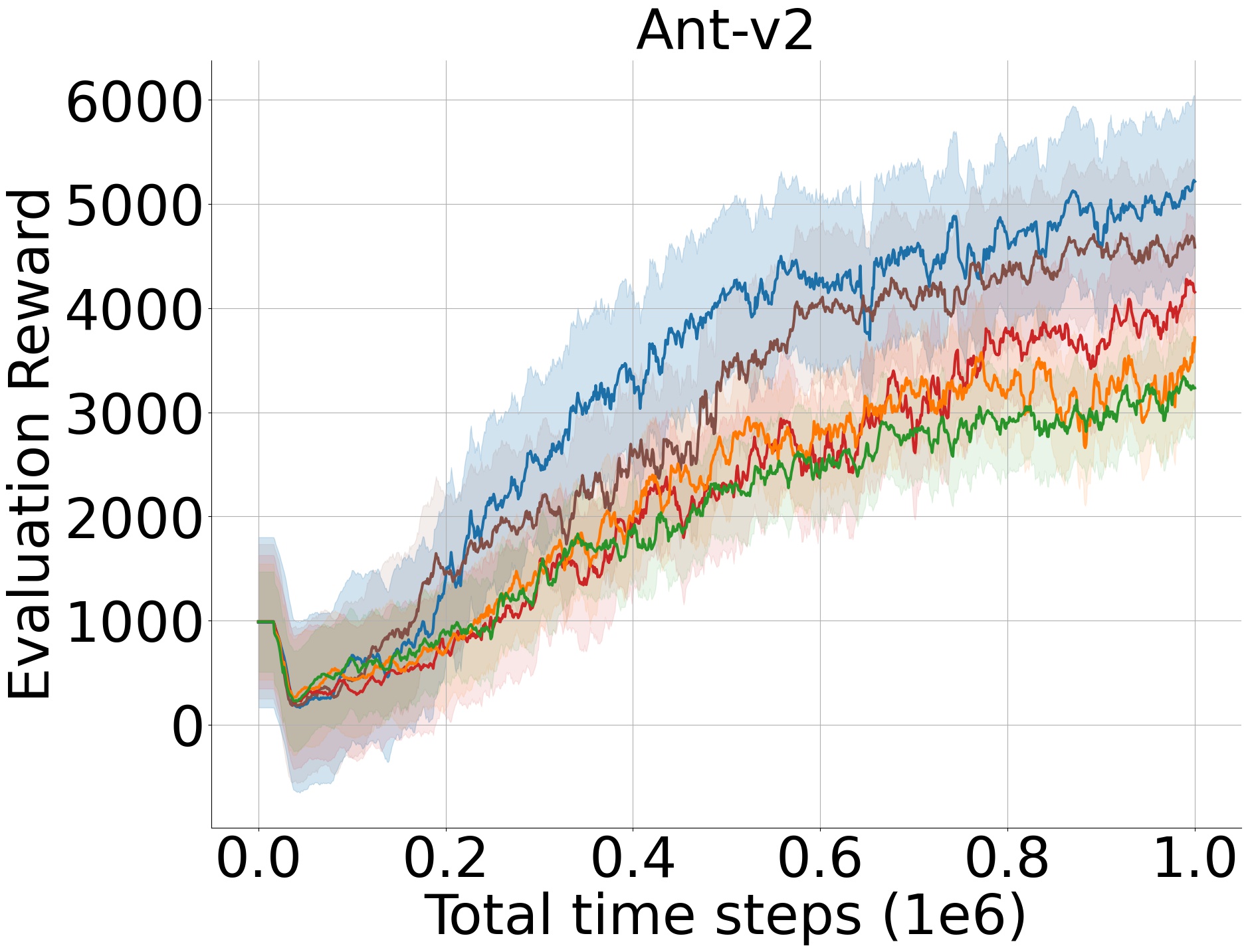}
		\includegraphics[width=1.6in, keepaspectratio]{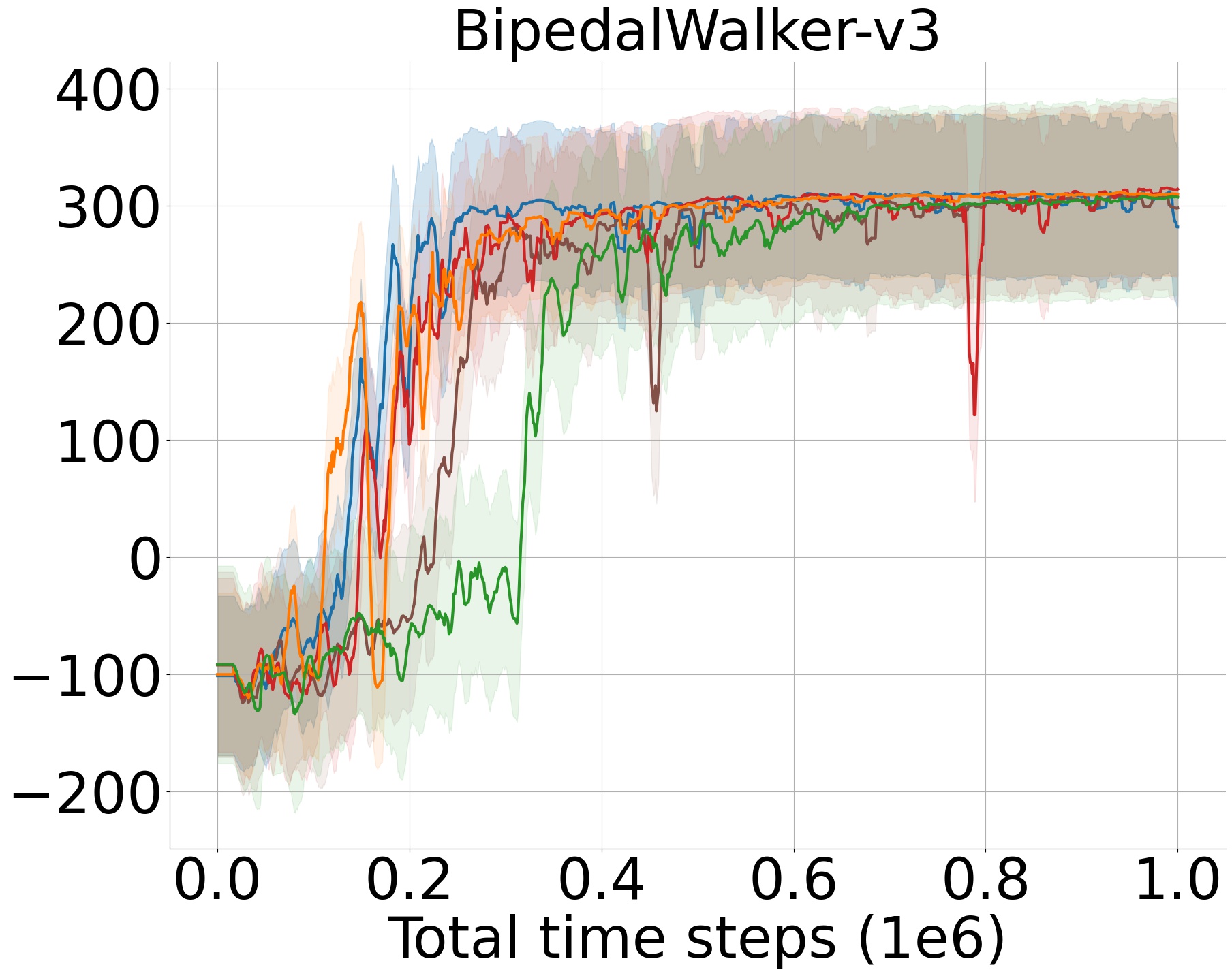}
		\includegraphics[width=1.6in, keepaspectratio]{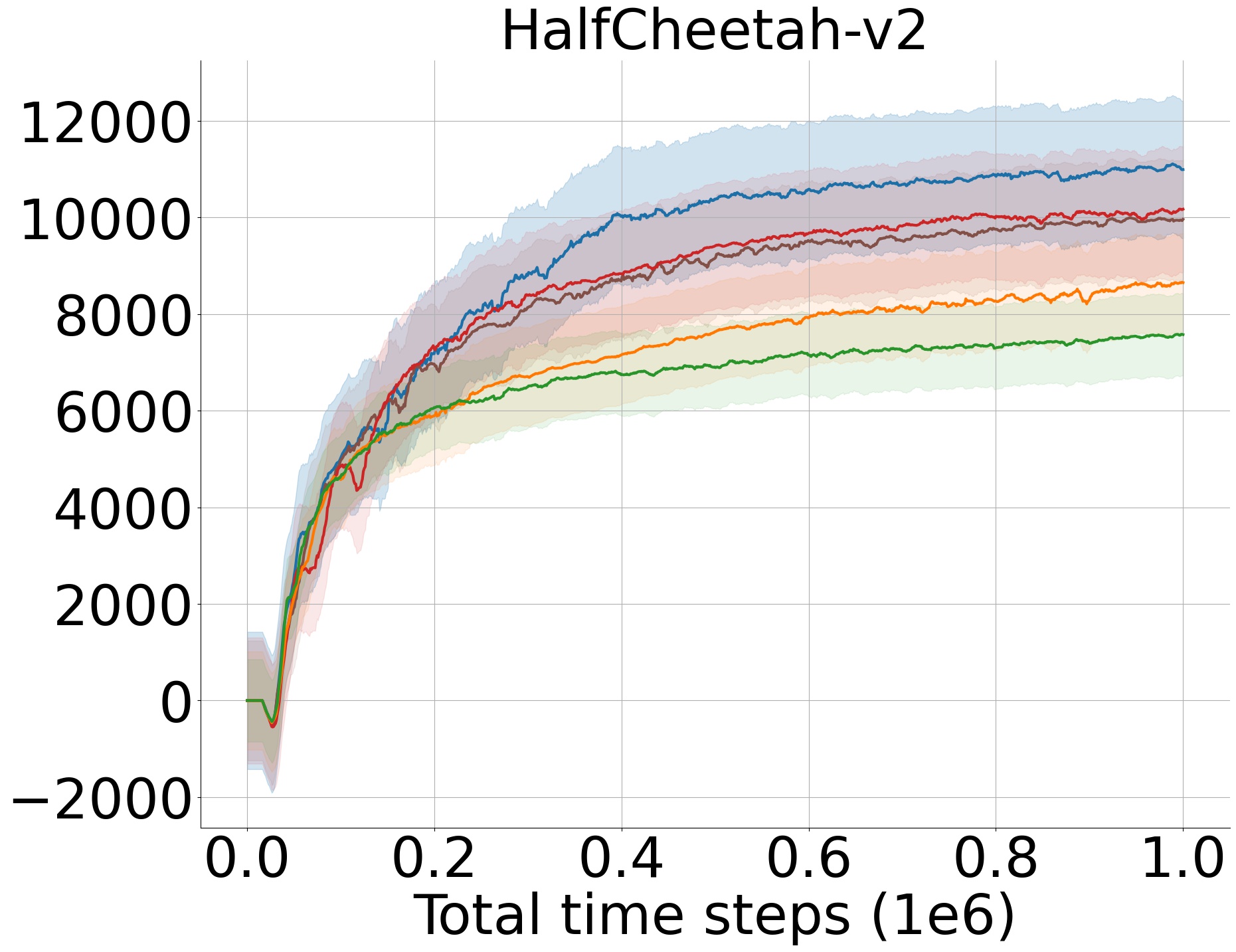}
		
	} \\
	\subfigure{
	    \includegraphics[width=1.6in, keepaspectratio]{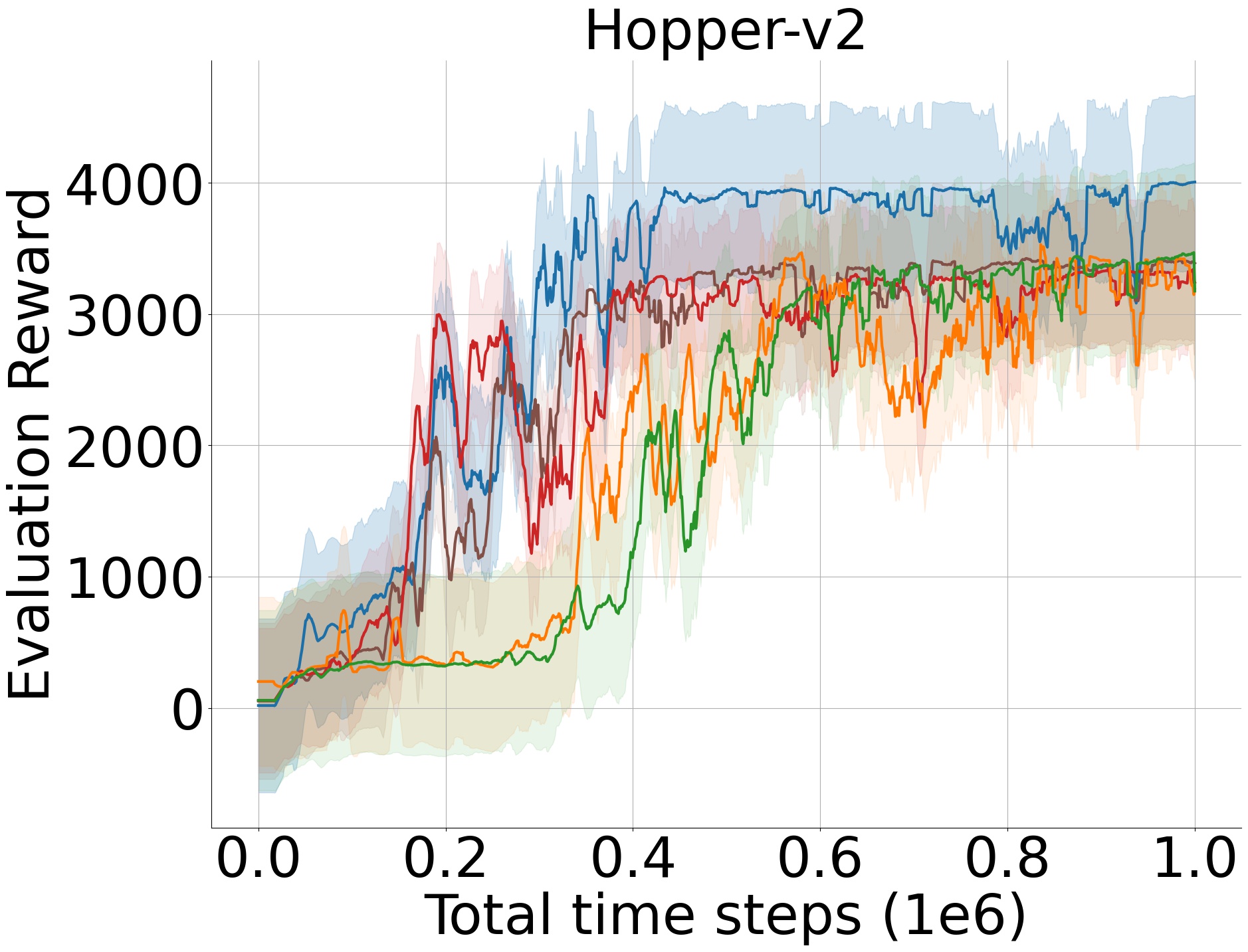}
		\includegraphics[width=1.6in, keepaspectratio]{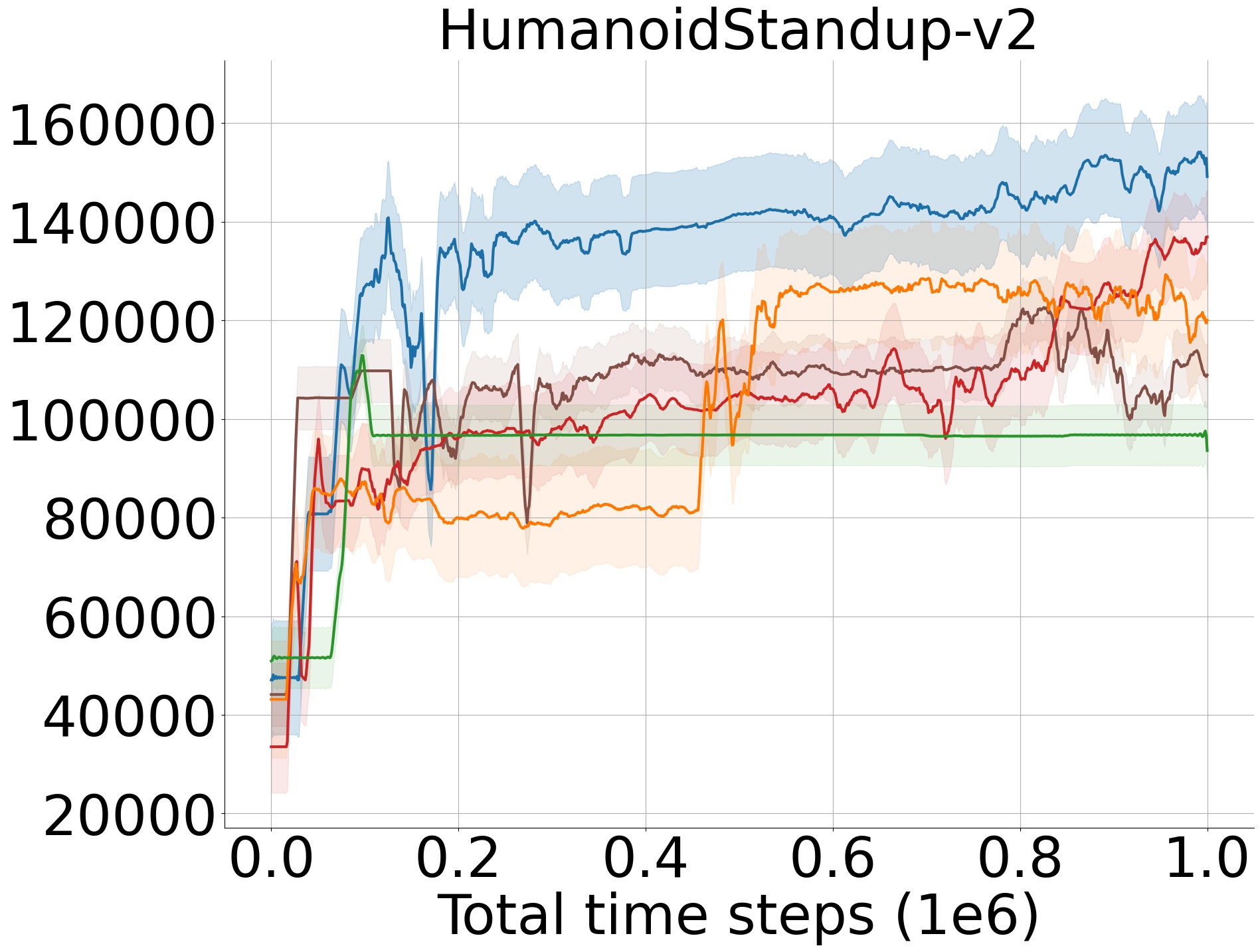}
		\includegraphics[width=1.6in, keepaspectratio]{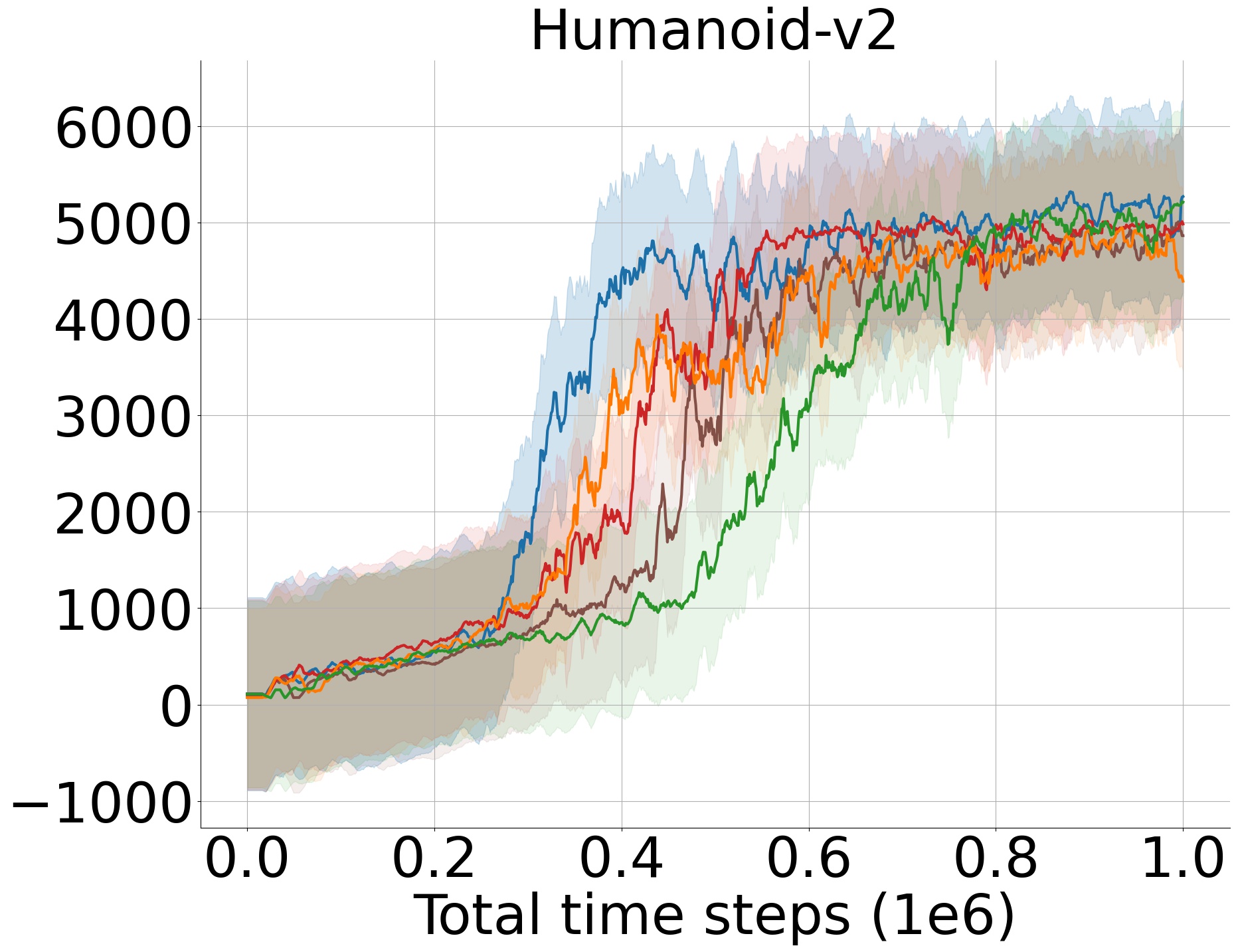}
	} \\
	\subfigure{
	    \includegraphics[width=1.6in, keepaspectratio]{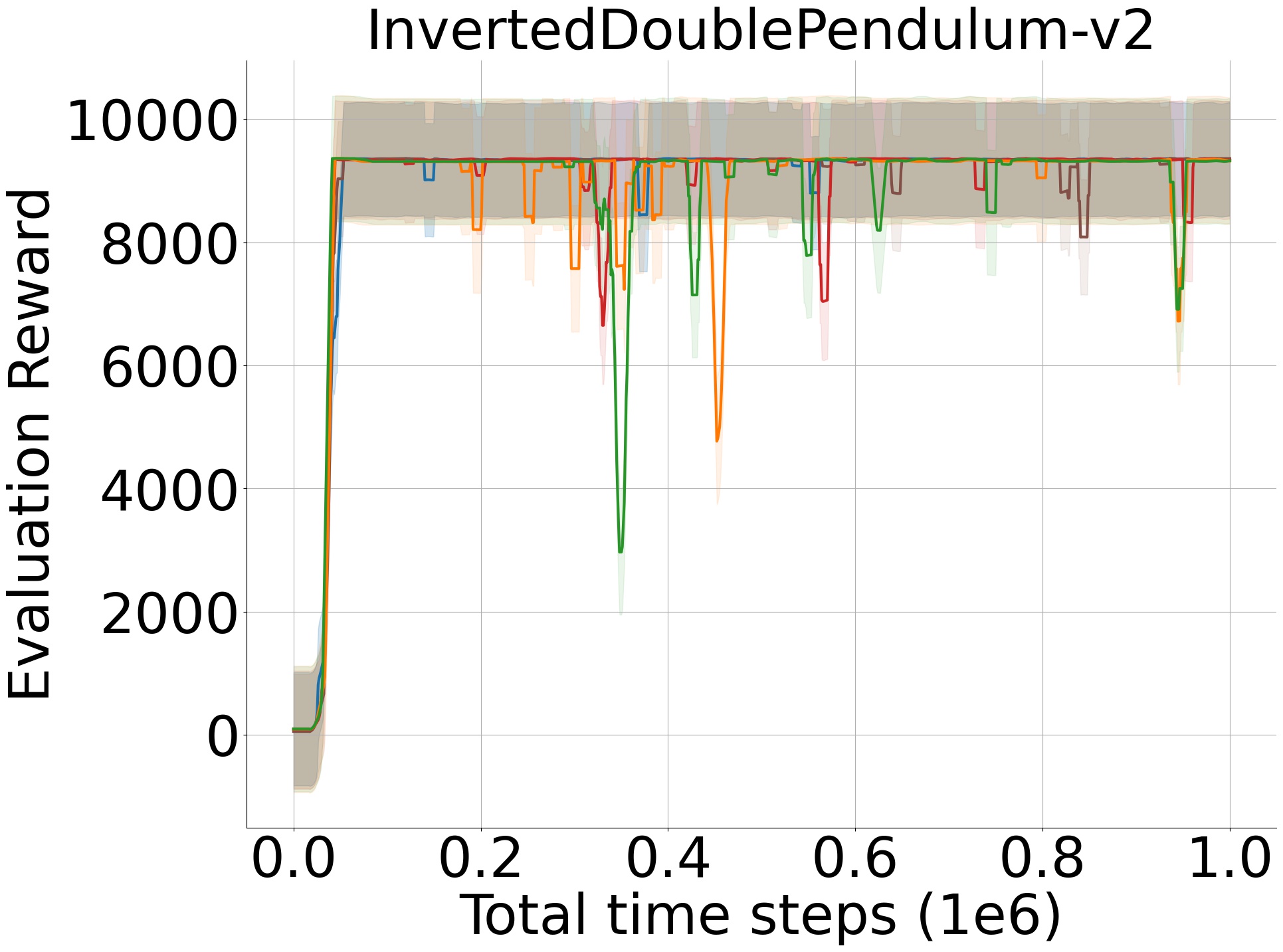}
		\includegraphics[width=1.6in, keepaspectratio]{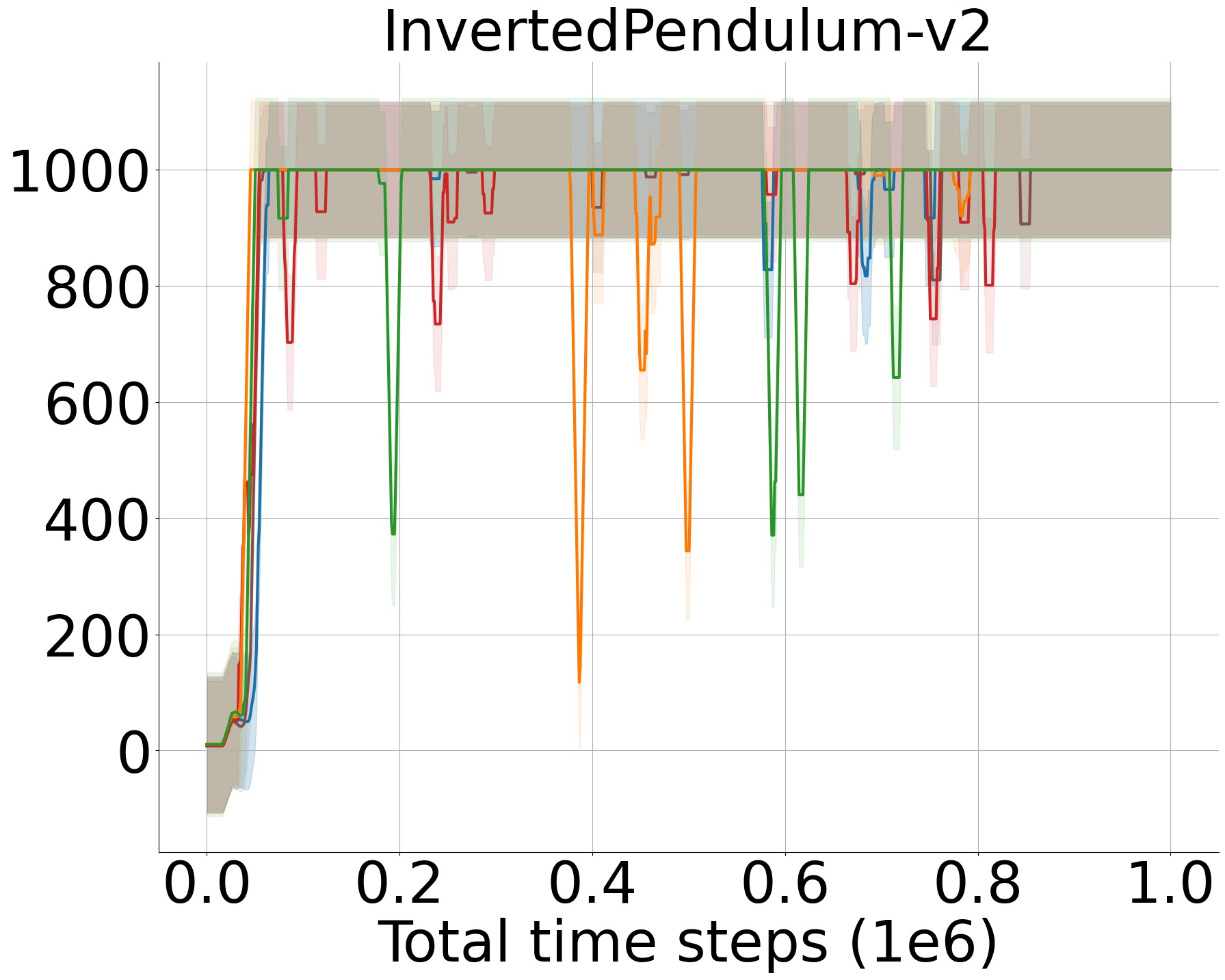}
		\includegraphics[width=1.6in, keepaspectratio]{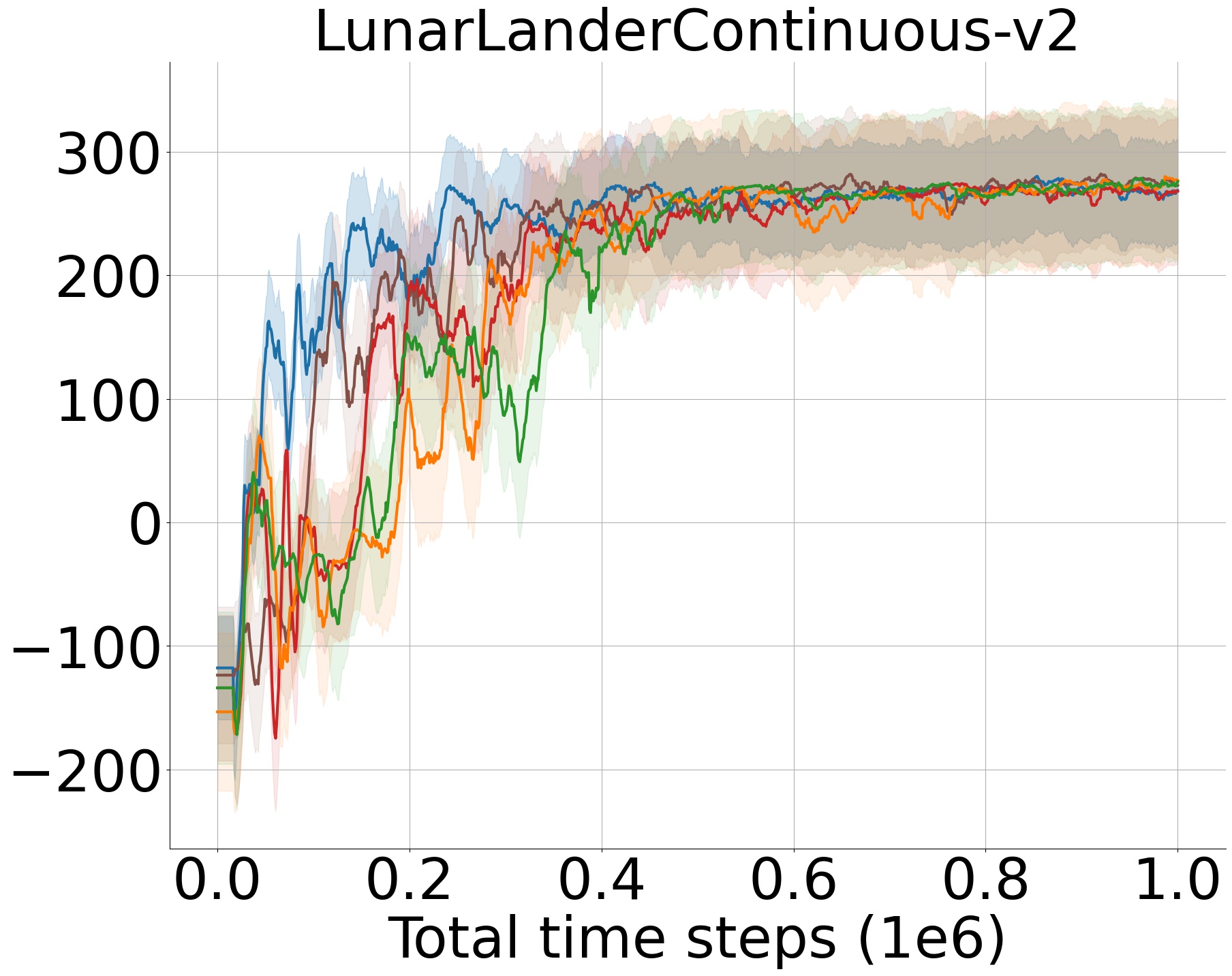}
	} \\
	\subfigure{
	    \includegraphics[width=1.6in, keepaspectratio]{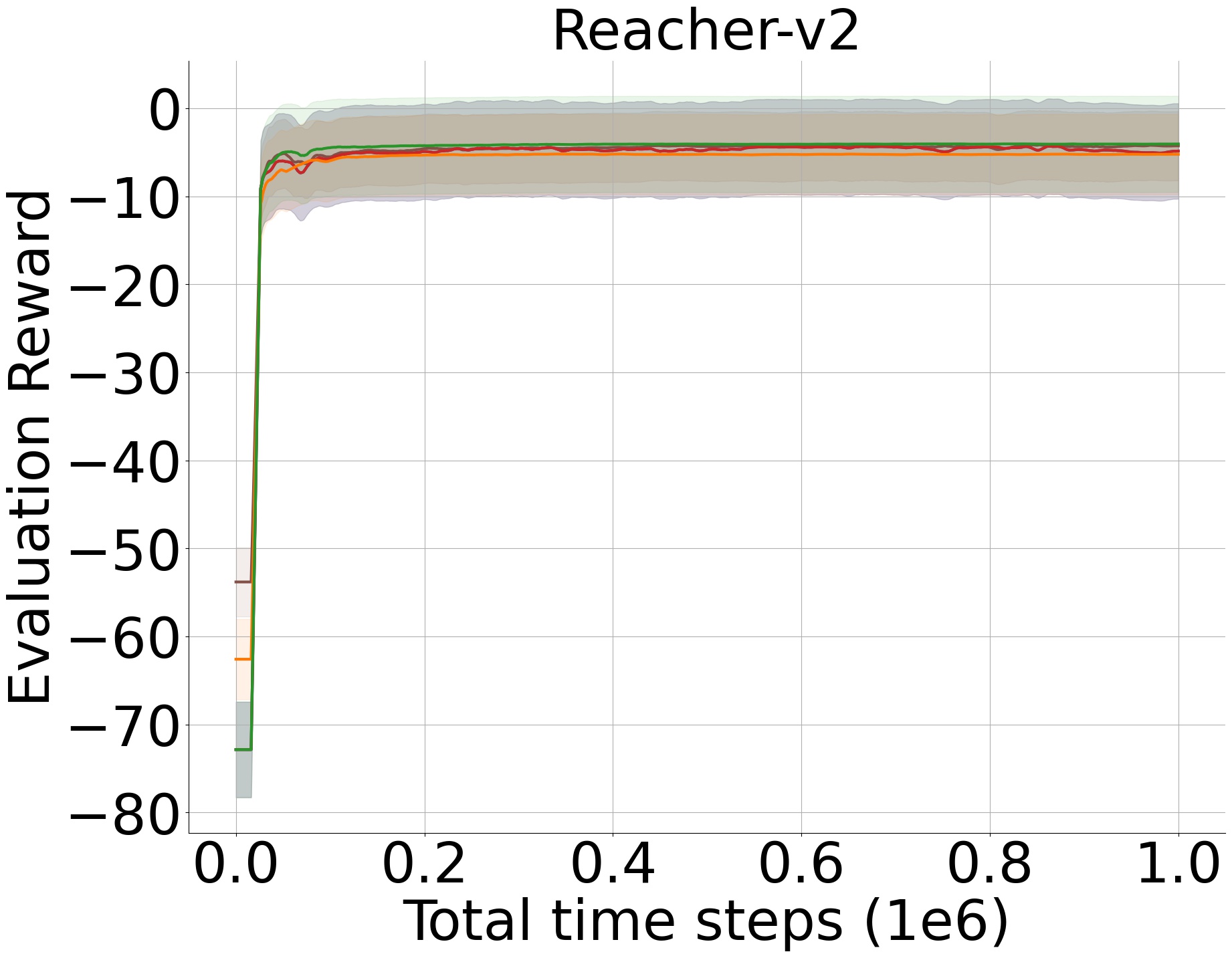}
		\includegraphics[width=1.6in, keepaspectratio]{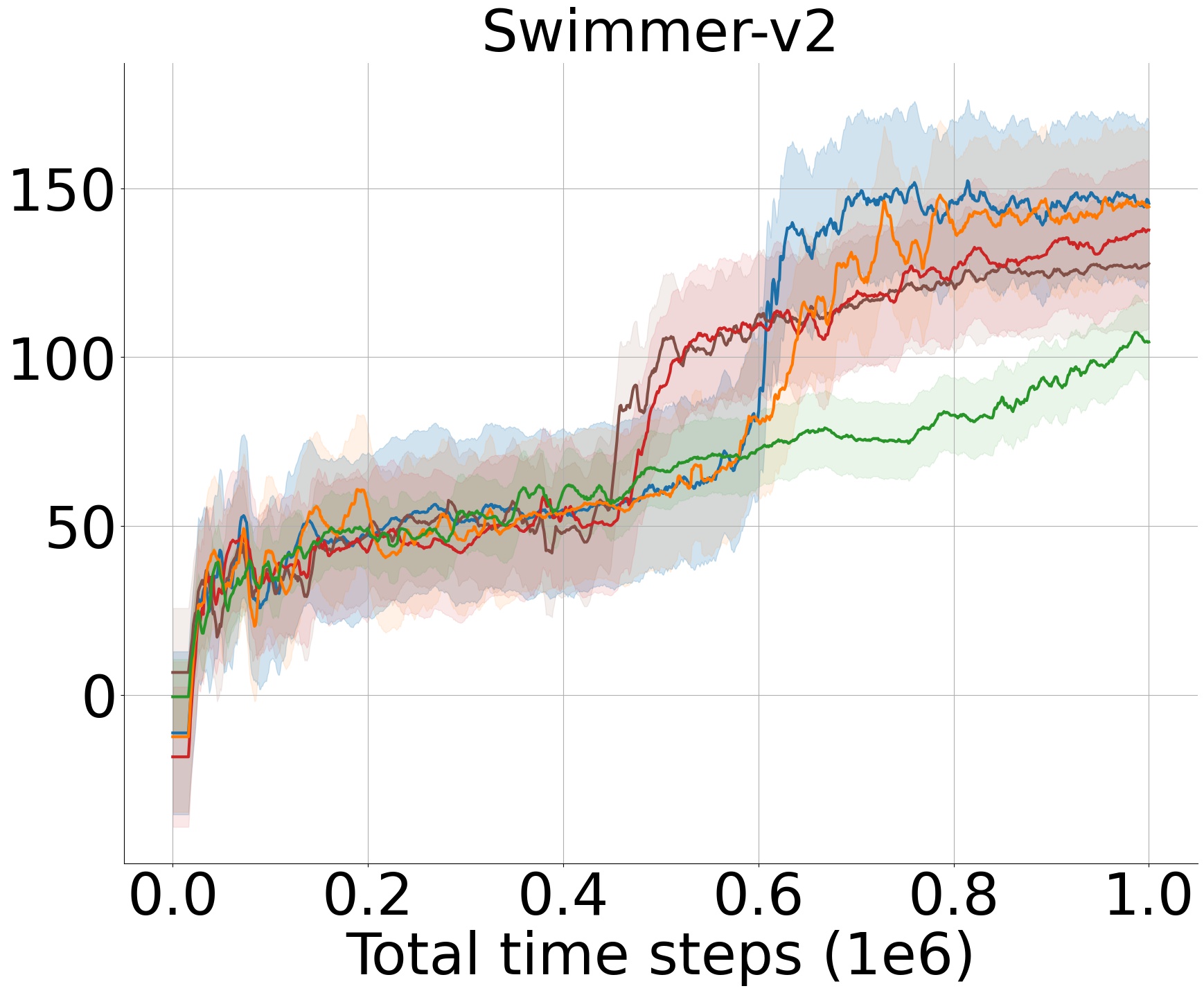}
		\includegraphics[width=1.6in, keepaspectratio]{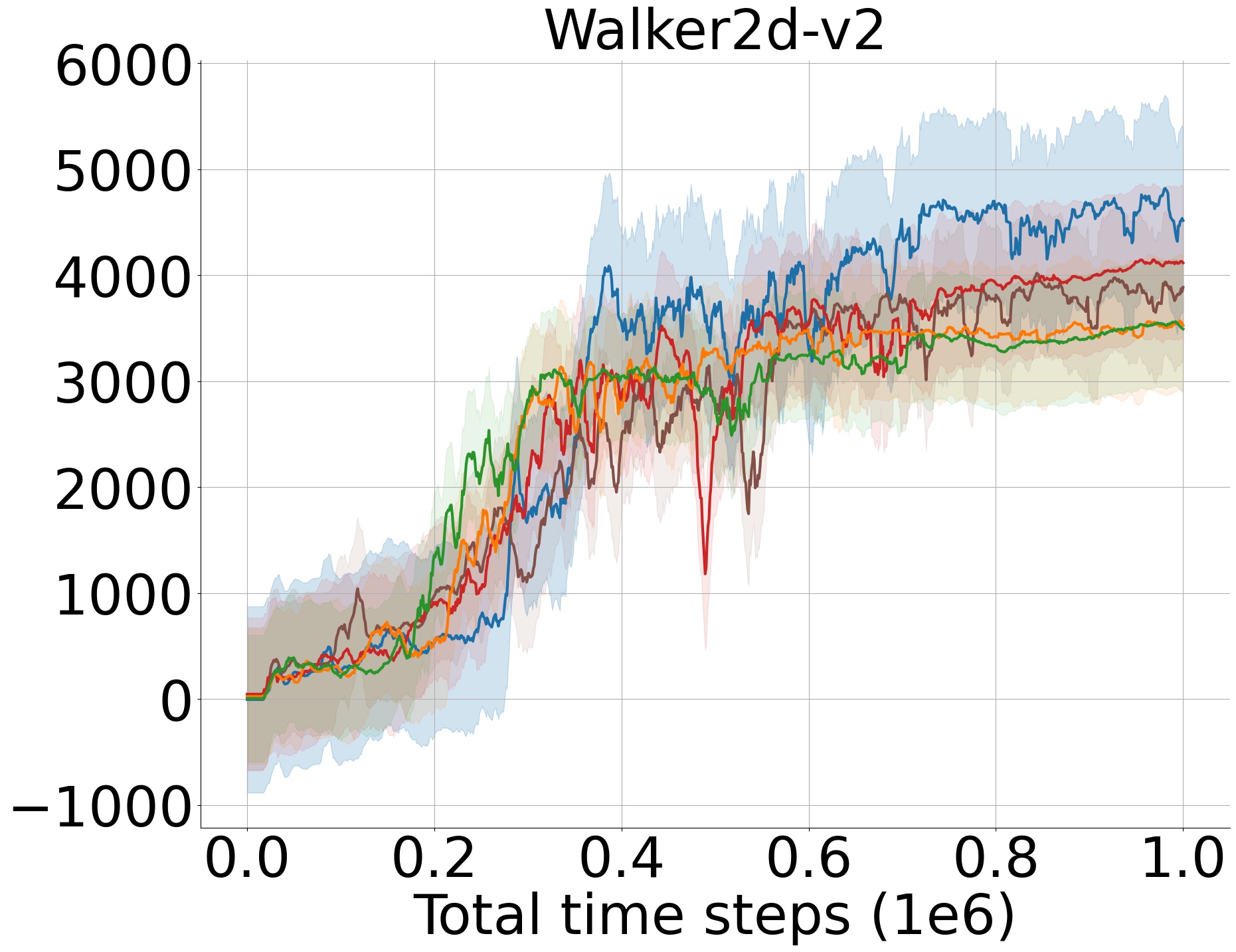}
	}
	\caption{Learning curves for the set of OpenAI Gym continuous control tasks. The shaded region represents half a standard deviation of the average evaluation over ten trials. Curves are smoothed uniformly with a sliding window of size 10.}
	\label{evaluation_results}
\end{figure*}

\subsection{Discussion}
\subsubsection{Q-value Comparisons}
Actual and estimated Q-value comparisons for our approach versus TD3 \cite{td3}, WD3 \cite{wd3} and TADD \cite{tadd} over six OpenAI Gym \cite{gym} continuous control tasks are reported in Fig. \ref{td3_swtd_q_estimation}, \ref{wd3_swtd_q_estimation} and \ref{tadd_swtd_q_estimation}, respectively. SWTD3 obtains more accurate Q-value estimates than TD3 \cite{td3} and the baseline algorithms in all of the environments tested. Our empirical findings indicate several cases. First, we observe in the baseline Q-value estimations that the underestimation increases since the variance of the received reward signals grows throughout the learning, reflecting Remark \ref{rem:increasing_variance}. Second, although our method obtains fairly accurate Q-value estimates and is not affected by an increasing reward variance, the Q-values are overestimated in the initial steps. This is due to the large $\beta$ values sampled at the beginning of the learning. However, as we discussed, such overestimated Q-values are tolerated by the agent, and the estimations reduce to a negligible margin of error, which verify our made claim in Remark \ref{rem:increasing_variance}. 

Furthermore, we fine-tune the $\beta$ value for the environments that are not reported in \cite{wd3} and \cite{tadd}, as stated previously. Our fine-tuning results show that the corresponding $\beta$ values in these environments are the same for WD3 \cite{wd3} and TADD \cite{tadd} since the expected function approximation error is also the same, as highlighted in Remark \ref{rem:wd3_tadd_same_est_error}. As a result, the mean estimation errors in these environments are practically the same for WD3 \cite{wd3} and TADD \cite{tadd}, particularly, BipedalWalker, HumanoidStandup, Humanoid, and LunarLanderContinuous. For the environments that are reported in \cite{wd3} and \cite{tadd}, WD3 \cite{wd3} obtains more accurate Q-value estimates than TADD \cite{tadd} since $\beta = 0.95$ used in the TADD algorithm \cite{tadd}, which corresponds to a significant underestimation error due to the large contribution of the negative reward variance, as specifically shown in (\ref{eq:tadd_est_error}). Our method attains substantially more accurate Q-value estimates than the competing approaches. It overcomes the effects induced by the increasing variance of the received reinforcement learning signals through sampling from an estimation error interval, the lower bound of which is constantly decreased, verifying Remark \ref{rem:variance_does_not_affect}. 

\begin{table*}[!hbt]
\scriptsize
\begin{center}
\caption{Average of last 10 evaluation returns over 10 trials. Boldface represents the maximum in each task. $\pm$ denotes the single standard deviation over trials. The WD3 and TADD algorithms use the beta values given in Table \ref{wd3_beta_values_table}.}
\label{eval_results}
    \begin{tabular}{@{} lccccc @{}}
        \toprule
        \textbf{Environment} & \textbf{SWTD3} & \textbf{TADD} & \textbf{WD3} & \textbf{TCD3} & \textbf{TD3} \\
        \midrule
        Ant-v2 & \textbf{5216.25 $\pm$ 156.85} & 3232.86 $\pm$ 316.71  & 4151.4 $\pm$ 259.28 & 4027.69 $\pm$ 326.34 & 3123.36 $\pm$ 306.44 \\ 
        BipedalWalker-v3 & \textbf{309.42 $\pm$ 1.02} & 297.95 $\pm$ 20.51  & 300.34 $\pm$ 8.93 & 307.27 $\pm$ 1.82 & 296.16 $\pm$ 32.75 \\
        HalfCheetah-v2 & 10990.44 $\pm$ 95.26 & 8651.46 $\pm$ 72.87  & 10170.53 $\pm$ 109.76 & 9961.2 $\pm$ 80.39 & 7574.84 $\pm$ 48.86 \\
        Hopper-v2 & \textbf{3655.96 $\pm$ 7.03} & 3438.8 $\pm$ 284.49  & 3242.01 $\pm$ 321.87 & 3387.83 $\pm$ 17.03 & 3175.41 $\pm$ 906.72 \\
        HumanoidStandup-v2 & \textbf{149164.28 $\pm$ 10951.44} & 120007.62 $\pm$ 7053.61  & 136939.36 $\pm$ 867.69 & 108915.83 $\pm$ 2534.29 & 93588.07 $\pm$ 10085.69 \\
        Humanoid-v2 & 5269.39 $\pm$ 154.65 & 4391.13 $\pm$ 470.96  & 4984.03 $\pm$ 122.46 & 4863.22 $\pm$ 266.93 & 5211.33 $\pm$ 129.84 \\
        InvertedDoublePendulum-v2 & \textbf{9357.99 $\pm$ 0.71} & 9350.29 $\pm$ 1.5  & 9353.49 $\pm$ 2.97 & 9334.67 $\pm$ 10.95 & 9350.29 $\pm$ 1.5 \\ 
        InvertedPendulum-v2 & \textbf{1000.0 $\pm$ 0.0} & \textbf{1000.0 $\pm$ 0.0}  & \textbf{1000.0 $\pm$ 0.0} & \textbf{1000.0 $\pm$ 0.0} & \textbf{1000.0 $\pm$ 0.0} \\
        LunarLanderContinuous-v2 & \textbf{277.11 $\pm$ 7.59} & 276.88 $\pm$ 4.58  & 268.1 $\pm$ 7.78 & 272.77 $\pm$ 5.21 & 275.85 $\pm$ 8.21 \\
        Reacher-v2 & \textbf{-3.53 $\pm$ 0.06} & -5.23 $\pm$ 0.03  & -4.86 $\pm$ 0.05 & -4.27 $\pm$ 0.03 & -4.06 $\pm$ 0.01 \\
        Swimmer-v2 & \textbf{145.54 $\pm$ 5.41} & 144.52 $\pm$ 4.81  & 137.64 $\pm$ 1.8 & 127.69 $\pm$ 2.24 & 104.5 $\pm$ 1.46 \\
        Walker2d-v2 & \textbf{4517.13 $\pm$ 287.1} & 3525.91 $\pm$ 94.07  & 4117.14 $\pm$ 67.13 & 3886.36 $\pm$ 221.19 & 3492.81 $\pm$ 34.7 \\
        \bottomrule
    \end{tabular}
\end{center}
\normalsize
\end{table*}

\subsubsection{Evaluation}
Table \ref{eval_results} reports the evaluation results in terms of the average of the last ten evaluation rewards over ten random seeds. Additionally, Fig. \ref{evaluation_results} depicts the corresponding learning curves. From our experimental results, we observe that our method either matches or outperforms the performance of TD3 \cite{td3} and baseline algorithms in terms of the learning speed and highest evaluation return. In the environments such as BipedalWalker, Humanoid, and LunarLanderContinuous, where our algorithm and competing approaches converge to the approximately same highest evaluation returns, Fig. \ref{evaluation_results} demonstrates that SWTD3 obtains a faster convergence by largely shrinking the underestimation bias and overcoming the increasing reward variance. Moreover, we do not observe a significant performance difference in trivial environments, e.g., InvertedDoublePendulum, InvertedPendulum, and Reacher, as they do not require complex solutions \cite{deep_rl_that_matters}. 

We observe that TCD3 \cite{tcd3}, WD3 \cite{wd3} and TADD \cite{tadd} exhibits a better performance than TD3 \cite{td3}. However, in the environments reported by \cite{tadd}, where $\beta = 0.95$, the performance of TADD \cite{tadd} is very similar to TD3 \cite{td3} as $\beta = 1.0$ corresponds to the same expected error in TD3 \cite{td3}. Furthermore, from our discussion in Remark \ref{rem:wd3_tadd_same_est_error} and \ref{rem:wd3_tadd_comp}, and theoretical analysis in (\ref{eq:expected_estimation_error_tcu}), (\ref{eq:wd3_est_error}) and (\ref{eq:tadd_est_error}), we infer that TCD3 \cite{tcd3}, WD3 \cite{wd3} and TADD \cite{tadd} yield approximately the same performance for $\beta = 0.5$, which is depicted in the BipedalWalker environment. In addition, when the $\beta$ value of WD3 \cite{wd3} is smaller than of TADD \cite{tadd}, it outperforms TADD \cite{tadd} since a small fixed $\beta$ value often corresponds to a decreased underestimation error. It exhibits the same performance in contrast, when the $\beta$ values are the same. Overall, these results are consistent with our Q-value comparisons, and reflect the theoretical insights made in this study. 

Ultimately, some methods exhibit a worse performance than outlined in the original articles. This is due to the stochasticity of the environment dynamics, that is, used dependencies, hardware, and random seeds have a large effect on the performance of reinforcement learning algorithms \cite{deep_rl_that_matters}. Nevertheless, we use the same set of seeds for all algorithms in our experiments, and evaluation results would be consistent if we used different seeds, which suffices a fair evaluation procedure \cite{deep_rl_that_matters}. This is also valid for the resulting performances when the same $\beta$ value is used for WD3 \cite{wd3} and TADD \cite{tadd}. The algorithmic differences alter the pseudorandom number order in the environment dynamics and cause the performances to differ slightly even under the same $\beta$ value. Nonetheless, the overall performances are practically the same.

\section{Conclusion}
In this paper, we focus on the underestimation of the Q-values in deterministic policy gradient \cite{dpg} methods. We extend our previous work on the underestimation by theoretically addressing the infeasible assumptions in the existing approaches that prevent them from adapting to off-policy actor-critic algorithms. We support our claims through Remarks and show that receiving different reward signals that vary on a large scale increases the underestimation of the action-value estimates. Then, through an extensive analysis of the estimation bias induced by the existing approaches, we introduce our novel Deep Q-learning \cite{dqn} variant that forms a linear combination of two Q-value approximators, with weights that are sampled from a shrunk estimation bias interval. Having our statistical analysis and extensive set of empirical studies combined, we demonstrate that the introduced approach notably outperforms the existing methods and improves our previous study. We also provide the exact implementation of the introduced algorithm at the GitHub repository for reproducibility concerns.

\ifCLASSOPTIONcaptionsoff
  \newpage
\fi

\bibliographystyle{IEEEtran}
\bibliography{references}

\end{document}